\documentclass{article}

\usepackage[dvipsnames]{xcolor}

\usepackage{microtype}
\usepackage{graphicx}
\usepackage{subcaption}
\usepackage{booktabs} 
\usepackage{amsfonts}
\usepackage{amsmath}
\usepackage{tabularx}
\usepackage{mathabx}
\usepackage{amssymb}

\usepackage{hyperref}

\usepackage[accepted]{icml2022}

\usepackage[algo2e,ruled,vlined]{algorithm2e}
\usepackage{dirtytalk}

\newcommand{\xxcomment}[4]{\textcolor{#1}{[$^{\textsc{#2}}_{\textsc{#3}}$ #4]}}
\newcommand{\nvg}[1]{\xxcomment{RubineRed}{N}{G}{#1}}
\newcommand{\bighat}[1]{\xxcomment{blue}{B}{H}{#1}}
\newcommand{\sds}[1]{\xxcomment{Plum}{S}{S}{#1}}
\newcommand{\agw}[1]{\xxcomment{red}{A}{W}{#1}}
\newcommand{\wjm}[1]{\xxcomment{green}{W}{M}{#1}}
\newcommand{\todo}[1]{\xxcomment{orange}{to}{do}{#1}}

\renewcommand{\nvg}[1]{}
\renewcommand{\bighat}[1]{}
\renewcommand{\sds}[1]{}
\renewcommand{\agw}[1]{}
\renewcommand{\wjm}[1]{}
\renewcommand{\todo}[1]{}

\renewcommand{\vec}[1]{\mathbf{#1}}

\newcommand{\SeqBase}{\mathcal{X}_{\mathrm{base}}}
\newcommand{\SeqPool}{\mathcal{X}_{\mathrm{pool}}}
\newcommand{\SeqCand}{\mathcal{X}_{\mathrm{cand}}}

\newcommand{\SeqBatch}{\mathcal{X}_{\mathrm{batch}}}

\icmltitlerunning{Accelerating Bayesian Optimization for Biological Sequence Design with Denoising Autoencoders}

\usepackage{appendix}

\begin{document}

\twocolumn[
\icmltitle{Accelerating Bayesian Optimization for Biological\\Sequence Design with Denoising Autoencoders}

\icmlsetsymbol{equal}{*}

\begin{icmlauthorlist}
\icmlauthor{Samuel Stanton}{nyu-cds}
\icmlauthor{Wesley J. Maddox}{nyu-cds}
\icmlauthor{Nate Gruver}{nyu-cs}
\icmlauthor{Phillip Maffettone}{big-hat}
\icmlauthor{Emily Delaney}{big-hat}\\
\icmlauthor{Peyton Greenside}{big-hat}
\icmlauthor{Andrew Gordon Wilson}{nyu-cds,nyu-cs}
\end{icmlauthorlist}

\icmlaffiliation{nyu-cds}{Center for Data Science, New York University, New York, USA}
\icmlaffiliation{nyu-cs}{Courant Institute of Mathematical Sciences, New York University, New York, USA}
\icmlaffiliation{big-hat}{BigHat Biosciences, San Mateo, CA, USA}

\icmlcorrespondingauthor{Samuel Stanton}{ss13641@nyu.edu}
\icmlcorrespondingauthor{Andrew Gordon Wilson}{andrewgw@cims.nyu.edu}

\icmlkeywords{Machine Learning, ICML}

\vskip 0.3in
]

\printAffiliationsAndNotice{} 
\begin{abstract} 
Bayesian optimization (BayesOpt) is a gold standard for query-efficient continuous optimization. However, its adoption for drug design has been hindered by the discrete, high-dimensional nature of the decision variables. 
We develop a new approach (LaMBO) which jointly trains a denoising autoencoder with a discriminative multi-task Gaussian process head, allowing gradient-based optimization of multi-objective acquisition functions in the latent space of the autoencoder.
These acquisition functions allow LaMBO to balance the explore-exploit tradeoff over multiple design rounds, and to balance objective tradeoffs by optimizing sequences at many different points on the Pareto frontier.   
We evaluate LaMBO on two small-molecule design tasks, and introduce new tasks optimizing \textit{in silico} and \emph{in vitro} properties of large-molecule fluorescent proteins.
In our experiments LaMBO outperforms genetic optimizers and does not require a large pretraining corpus, demonstrating that BayesOpt is practical and effective for biological sequence design. 
\end{abstract}

\section{Introduction}
\label{sec:intro}

Modern drug development is a very costly endeavor, with estimates ranging from \$310M to \$2.8B for each new drug \citep{wouters2020estimated}.
There are three major phases of drug development: 1) \textit{target discovery}, the proposal of a biological mechanism hypothesized to treat a specific medical condition, 2) \textit{drug design}, the specification of a molecular payload which will interact with the proposed mechanism, and 3) \textit{clinical trials}, the evaluation of the efficacy and safety of the payload \textit{in vivo}.
Though each phase contributes to the total cost of development, in this work we focus on the drug design phase.
In particular we will optimize real-valued target properties (such as neurotransmitter receptor affinity or protein folding stability) for payloads represented as discrete sequences (e.g. SMILES strings).
Since the mappings from sequence to target are often unknown, expensive to observe \textit{in vitro}, and difficult to simulate, we pose biological sequence design as a costly black-box optimization (BBO) problem over a large discrete search space.

\begin{figure}[t!]
    \centering
    \hspace{-4mm}
    \includegraphics[width=0.95\linewidth,trim=0.5cm 0.9cm 0.5cm 0.75cm,clip]{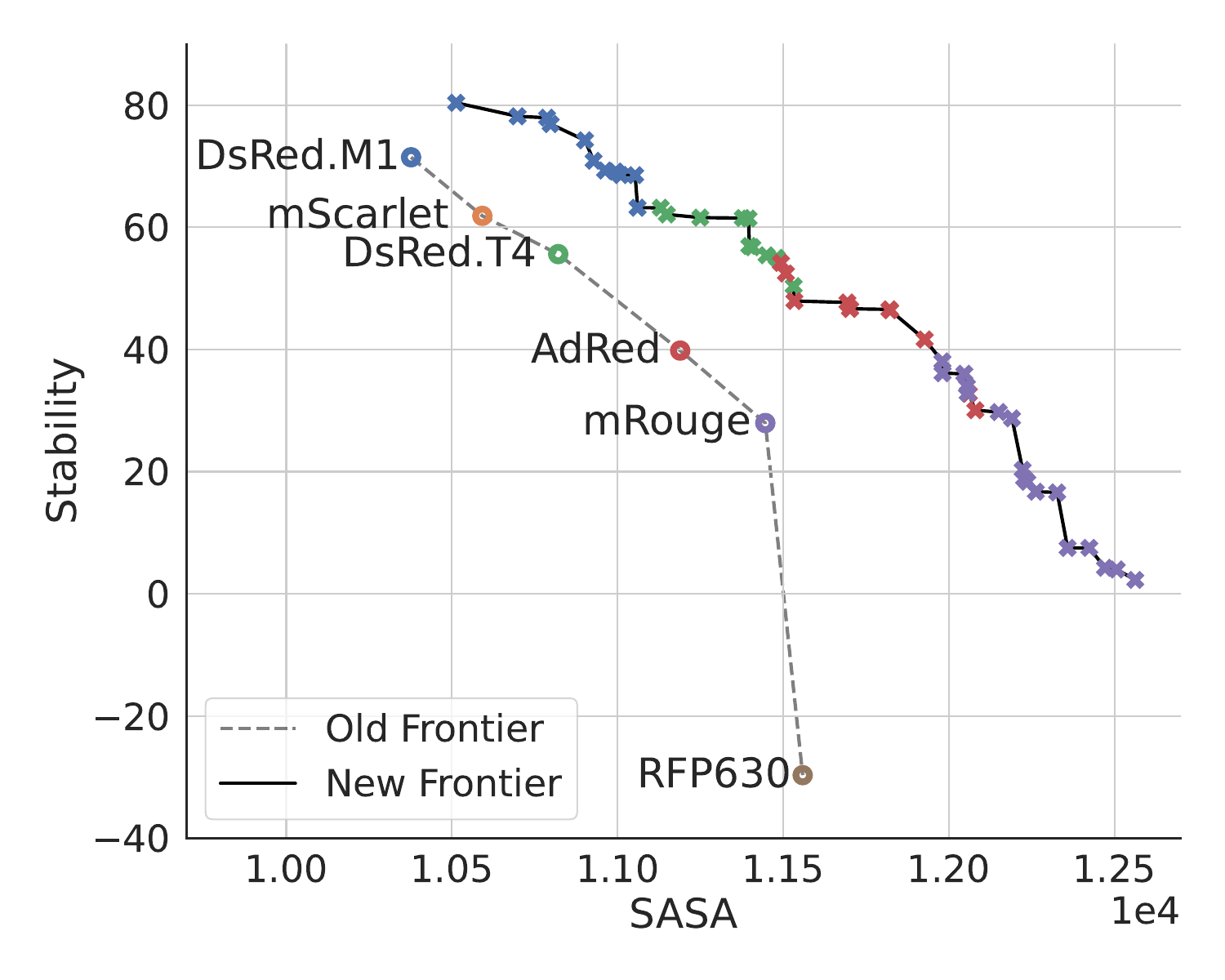}
    \vspace{-0.25cm}
    \caption{BayesOpt can be used to maximize the simulated folding stability (i.e. -dG, or the negative change in Gibbs free energy) and solvent-accessible surface area (SASA) of red-spectrum fluorescent proteins (RFPs). Higher is better for both objectives. The ancestor proteins are shown as colored circles, with corresponding optimized offspring shown as crosses. Stability correlates with protein function (e.g. how long the protein can fluoresce) while SASA is a proxy for fluorescent intensity.}
    \label{fig:intro_pareto_frontier}
\end{figure}

Recent work applying deep learning to biological tasks has primarily focused on learning from a static, offline dataset \citep{rao2019evaluating, jumper2021highly, baek2021accurate, meier2021language, rives2021biological}.
When these models are used to select new sequences to label, they are applied in a one-shot fashion, without accounting for future design rounds \citep{gligorijevic2021function, biswas2021low}.
Because labels are scarce for many important targets, it is important to account for model uncertainty to manage the explore-exploit tradeoff \citep{o2018uncertainty}.

Bayesian optimization (BayesOpt) is a powerful class of BBO algorithms, explicitly designed to \textit{coherently} reason about online decision-making based on incomplete information \citep{brochu2010tutorial}.
BayesOpt balances the explore-exploit tradeoff in a principled way, relying on a probabilistic discriminative model to prioritize decisions with the highest potential payoff.
At each decision point the discriminative model produces a posterior distribution over the hypothesis space of all functions the model can represent. 
The posterior formally represents the degree to which any particular hypothesis is plausible given the data and model assumptions.
To make a decision, BayesOpt selects the best decision as defined by an \emph{acquisition function}, such as expected improvement (EI), with each hypothesis contributing to the acquisition value in proportion to its posterior probability \citep{jones1998efficient}.
BayesOpt is not only philosophically appealing, it is provably a \textit{no-regret} strategy under the right conditions \citep{srinivas2010gaussian}. 

Like many Bayesian methods, the barriers hindering widespread adoption of BayesOpt are not conceptual, but \textit{practical}.
Drug design in particular is a natural application domain, but introduces multiple challenges. 
\textit{Discrete, high-dimensional inputs:} antibody therapeutic payloads are often RNA proteins which instruct the patient's immune system to produce the desired antibody. 
When proteins are represented as a sequence of residues, each identifying one of 20 possible amino acids, even a fairly small 200-residue protein is only one of $20^{200} \approx 1.6 \times 10^{260}$ options.

By contrast, conventional BayesOpt works best on problems with 10 or fewer continuous decision variables, due to the properties of standard kernels, and the lack of gradients to maximize the acquisition function.
\textit{Batched, multi-objective experiments:} because considerations including efficacy, toxicity, and yield must all be taken into account, drug design is inherently multi-objective. 
Furthermore sequences are labeled in batches, necessitating the use of more sophisticated acquisition functions than standard workhorses like EI.
\textit{Data-scarcity:} wet lab experimental data is expensive and challenging to collect, so it is rare to have large-scale datasets with labels for the exact target properties of interest.

\begin{figure}[t]
    \centering
    \includegraphics[width=0.4\textwidth]{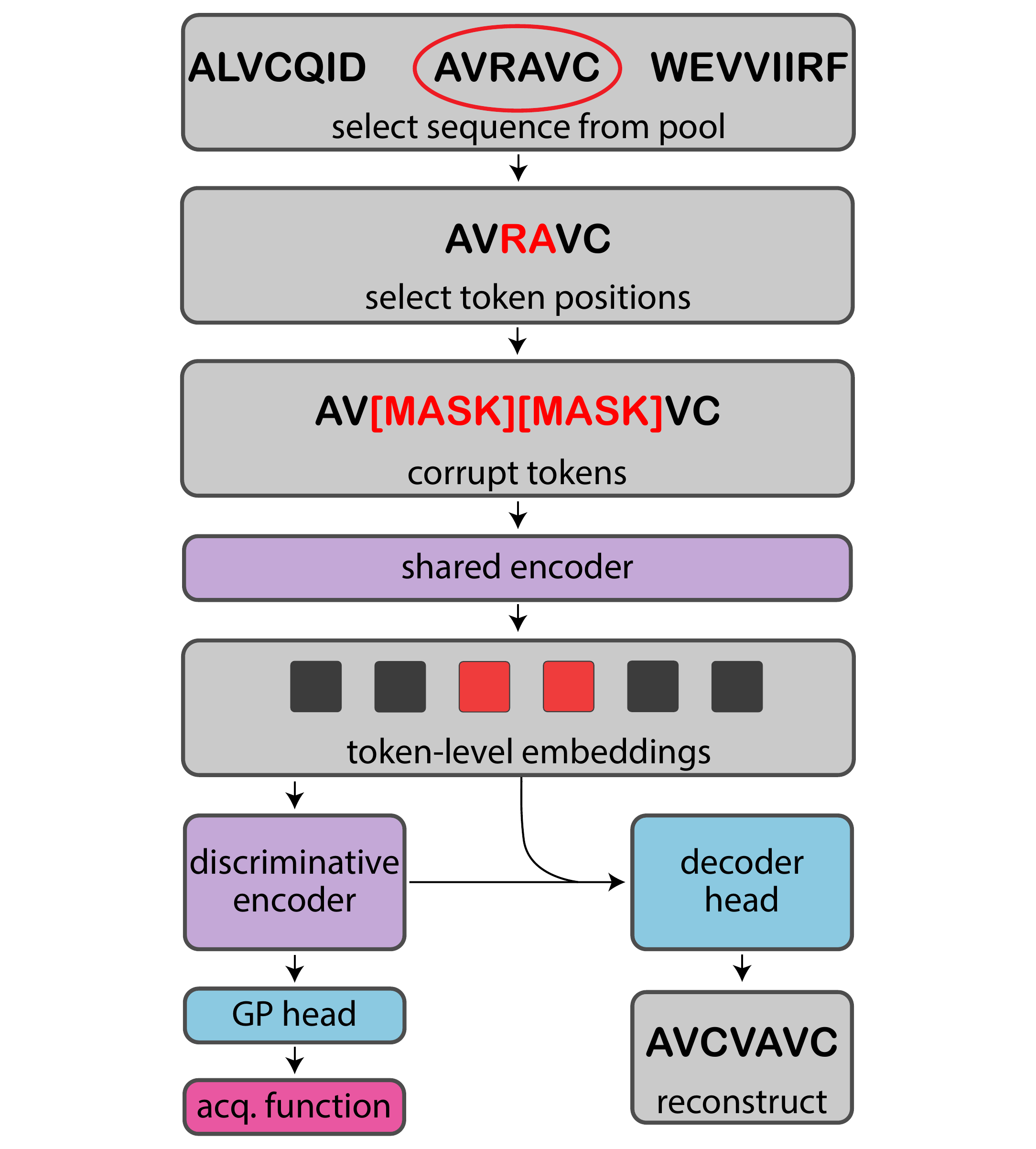}
    \caption{
    LaMBO with a non-autoregressive denoising autoencoder (DAE) architecture:
    a shared encoder produces continuous token-level embeddings $Z$ from corrupted inputs, which are passed to a discriminative encoder to produce target-specific token-level embeddings $Z'$. 
    A generative decoder head receives both $Z$ and $Z'$ as input, and a discriminative Gaussian process (GP) head pools $Z'$ to predict the objective values.
    The GP head allows the use of principled acquisition functions to manage the explore-exploit tradeoff, and the DAE head allows LaMBO to follow the acquisition gradient in latent space when selecting new queries.
    }
    \label{fig:optimizer_diagram}
\end{figure}
\vspace{-2mm}

No single BayesOpt method simultaneously addresses all of these challenges (see Section \ref{sec:related_work}).
Thus, previous methods have necessarily only been evaluated on very stylized tasks that fail to capture key aspects of real drug design problems.
In this work we present \textbf{La}tent \textbf{M}ulti-Objective \textbf{B}ayes\textbf{O}pt (\textbf{LaMBO}) to address this deficiency, and propose a novel \textit{in silico} task which emulates protein design tasks more closely than common open-source drug design benchmarks.

We preview our results applying LaMBO to this new task in Figure \ref{fig:intro_pareto_frontier}, maximizing the stability and SASA of RFPs derived from the fpBase dataset \citep{lambert2019fpbase}, with the initial Pareto frontier in objective space shown as a dashed line.
The new Pareto frontier (the solid line) discovered by LaMBO is superior, as it is characterized more densely by new sequences that are Pareto improvements over their ancestors.
In short, our contributions are as follows:
\begin{enumerate}
    \item We propose the LaMBO architecture, a novel combination of a generative DAE with a discriminative deep kernel learning GP, with a simple joint training procedure and an effective decision optimization routine.
    \item We propose a new \textit{in silico} large-molecule task to augment existing open-source drug design benchmarks.
    \item We evaluate LaMBO \textit{in silico} on two small-molecule tasks and our new large-molecule task, comparing to both genetic and latent-space BBO baselines, showing improved sample efficiency and solution quality.\footnote{Code: \url{github.com/samuelstanton/lambo}.}
    \item We present \textit{in vitro} wet lab results using LaMBO to discover brighter, more thermostable red fluorescent proteins.
\end{enumerate}

\section{Preliminaries}

We now introduce the problem setting and BayesOpt.

\subsection{Discrete Multi-Objective Sequence Design}

We first define the input space $\mathcal{X} = \bigtimes_{i=1}^t \mathcal{V}$, 
where $\mathcal{V}$ is an ordered, discrete vocabulary of $v$ tokens, $t$ is the max sequence length, and $\bigtimes$ is the Cartesian product. $\mathcal{V}$ includes a padding token to accommodate sequences of varying length.
Because $|\mathcal{X}| = |\mathcal{V}|^t$ is exponential in $t$, $\mathcal{X}$ becomes too large to enumerate quickly as the sequence length grows, even if $|\mathcal{V}|$ is relatively small.
As a result, sequence optimization usually starts with a library or pool of initial sequences (see Figure \ref{fig:optimizer_diagram}, top), which are modified to produce new candidate sequences.
When posed in this way, the optimization problem is restructured into three nested decisions: (1) Choose a base sequence from the pool. (2) Choose which positions on the sequence to change. (3) Choose how the sequence will be changed at those positions.

We represent sequence design as a multi-objective optimization problem $\min_{\vec x \in \mathcal{X}}(f_1(\vec x), \dots, f_k(\vec x))$, where $k$ is the number of objectives and each $f_i: \mathcal{X} \rightarrow \mathbb{R}$ is an expensive, black-box function of decision variables $\vec x \in \mathcal{X}$.
Given two feasible solutions $\vec x$ and $\vec x'$, $\vec x$ \textit{dominates} $\vec x'$ ($\vec x \prec \vec x'$) if $f_i(\vec x) \leq f_i(\vec x') \; \forall i \in \{1, \dots, k\}$, and $\exists i \in \{1, \dots, k\}$ s.t. $f_i(\vec x) < f_i(\vec x')$.
In general, there will not be a single dominating solution $\vec x^*$; instead, we define the set of non-dominated solutions (i.e. the true \emph{Pareto frontier} $\mathcal{P}^*$),
\begin{align}
    \mathcal{P}^* := \{ \vec x \in \mathcal{X} \; | \; 
    \{ \vec x' \in \mathcal{X} \; | \; \vec x' \prec \vec x, \vec x' \neq \vec x \} = \emptyset\}.
\end{align}
Since $\mathcal{P}^*$ is unknown, we seek a set of candidate solutions $\mathcal{P}$ that are close in objective space to those in $\mathcal{P}^*$.
We find these solutions by maximizing the hypervolume bounded by the extremal points in $\mathcal{P} \cup \{\vec x_{\mathrm{ref}}\}$, where $\vec x_{\mathrm{ref}}$ is some reference solution.

\subsection{BayesOpt}

See \citet{brochu2010tutorial} and \citet{frazier2018tutorial} for a more complete introduction to BayesOpt.
BayesOpt constructs a probabilistic \emph{surrogate model} $\hat f \in \mathcal{F}$ which is trained to emulate $f$ from a dataset $\mathcal{D}_n := \{(\vec x_1, \vec y_1), \dots, (\vec x_n, \vec y_n) \}$, where $\vec y_i$ are noisy observations of $f(\vec x_i)$ (e.g. $\vec y_i = f(\vec x_i) + \varepsilon_i, \; \varepsilon_i \sim \mathcal{N}(\vec 0, \sigma^2I_k)$).
The surrogate posterior distribution is used to define an \emph{acquisition function} $a: \mathcal{X} \times \mathcal{F} \rightarrow \mathbb{R}$, which in turn defines an \textit{inner loop} optimization problem to select new query point(s).
The objective function is then queried at the selected points and the surrogate is retrained on the augmented dataset, and the procedure repeats, forming an \textit{outer loop} (Algorithm \ref{alg:bayes_opt_outer}).

GPs \citep{rasmussen2006gaussian} are often preferred as surrogates because they have closed-form posteriors and work well in data-scarce regimes.
The inductive biases of a GP are primarily determined by the choice of kernel, which defines a prior distribution over the values of $f$ for any finite collection of inputs.
Most commonly used GP kernels (e.g. RBF or Mat\'ern) rely on $\ell_2$ distance between inputs to determine the prior covariance between outputs. 
When the inputs are low-dimensional (e.g. $d=10$) such kernels work well, but in high dimensions the $\ell_2$ norm is often a poor choice of distance metric \citep{srinivas2010gaussian,wang2016bayesian}.
This limitation has motivated the development of \emph{deep kernel learning} (DKL), which learns a low-dimensional continuous embedding via an encoder such as a convolutional neural network (CNN) \citep{wilson2016deep}.
Although GPs are kernel-based models, there is a range of well-known methods to scale them to large, online datasets, notably inducing point methods like stochastic variational GPs (SVGPs) which admit the use of stochastic optimizers \citep{hensman2013gaussian, wilson2016stochastic, maddox2021conditioning}.

\begin{algorithm}[t]
\SetAlgoLined
\textbf{Inputs: } \text{hypothesis space } $\mathcal{F}$, \text{ acquisition } $a$, \text{dataset } $\mathcal{D}_0$. \\
\For{$i = 0, \dots, i_{\max} - 1$}{
    Fit $\hat f \in \mathcal{F}$ to $\mathcal{D}_i$. \\
    $\vec x^*_i = \min_{\vec x \in \mathcal{X}} a(\vec x, \hat f)$. $\; \leftarrow$ the inner loop \\
    Observe $\vec y_i \sim p(\cdot | \vec x^*_i)$. \\
    $\mathcal{D}_{i+1} = \mathcal{D}_i \cup (\vec x^*_i, \vec y_i)$. \\
    $\mathcal{P}_{i + 1} = \mathrm{nondominated}(\mathcal{D}_{i + 1})$.
}
\Return $\mathcal{P}_{i_\mathrm{max}}$
\caption{The BayesOpt outer loop}
\label{alg:bayes_opt_outer}
\end{algorithm}

\section{Related Work}
\label{sec:related_work}

\textbf{Discrete Optimization by Sampling:} genetic algorithms (GAs) such as NSGA-II \citep{deb2002fast} slowly evolve a good solution by random mutation. GAs are a simple, popular baseline for BBO problems, but are known for being inefficient \citep{turner2021bayesian}.
One solution is to continue generating mutations randomly, but screen the proposed queries with a discriminative model before labeling \citep{nigam2019augmenting, yang2019machine}.
Other solutions focus on proposing mutations more intelligently, including RL-based approaches \citep{angermueller2020population, Angermueller2020Model-based} and generative approaches \citep{jensen2019graph,biswas2021low, zhang2021unifying}.
In particular the generative approach described by \citet{lee2018deterministic} and \citet{gligorijevic2021function} inspired the LaMBO architecture.
All the approaches just discussed are greedy in the sense that they rely on point estimates of the objective values to select new queries.

\textbf{Discrete BayesOpt:} excluding library-based approaches such as \citet{yang2019batched}, discrete BayesOpt methods can be categorized by how they structure the inner loop optimization problem.
Some methods use substring kernels (SSKs), optimizing queries directly in sequence space with a GA, and evaluating only on small tasks \citep{lodhi2002text, beck2017learning, moss2020boss}.
\citet{khan2022antbo} is an example of concurrent antibody design work in this vein, exploiting task-specific knowledge of complementarity-determining regions (CDRs) of the antibodies to make the problem tractable.
See Appendix \ref{subsec:ssk_offline_regression} for more discussion and an experiment comparing SSK and DKL GPs in the offline regression setting.

Latent-space optimizers learn continuous sequence embeddings $Z$, which are shared by a generative decoder modeling $p(\vec x | Z)$ and the discriminative surrogate modeling $p(\vec y | Z)$ \citep{deshwal2021combining, grosnit2021high, maus2022local}.
Thus $Z$ can be optimized with gradient-based methods to produce new sequences.
LaMBO is most similar to Latent-Space BayesOpt (LSBO) \citep{gomez2018automatic}, since both methods model $p(\vec y | Z)$ as a GP and model $p(\vec x | Z)$ via an autoencoder.
LSBO uses a VAE pretrained on a large dataset to solve single-objective tasks, training the VAE weights and the auxiliary GP head separately.
With a specialized architecture proposed by \citet{jin2018junction} and a biased VAE objective proposed by \citet{tripp2020sample}, LSBO has been shown capable of solving simple small-molecule tasks such as maximizing penalized logP.
Aside from LaMBO's use of DAEs rather than VAEs, this work improves upon LSBO in multiple ways, such as enabling the use of a general-purpose architecture for both small and large molecules, removing the need to pretrain the autoencoder, providing a reliable procedure for jointly training generative and GP heads with a shared encoder, and the introduction of multi-objective tasks and acquisition functions.
See Appendix \ref{subsec:lsbo_comparison} for a comparison between LSBO and LaMBO in the single-objective BBO setting.

\textbf{Multi-Objective BayesOpt:}
\citet{daulton_differentiable_2020} proposed a batch version of expected hypervolume improvement (EHVI) \citep{emmerich2005single,emmerich2011hypervolume}, an extension of EI to multiple objectives.
In follow-up work, \citet{daulton2021parallel} proposed the noisy expected hypervolume improvement (NEHVI) acquisition function, which extends noisy expected improvement (NEI) \citep{letham2019constrained} to multiple objectives.
Multi-task GPs (MTGPs) have previously been used as surrogates for multi-objective BayesOpt \citep{shah2016pareto}, including recent work scaling MTGPs up to thousands of training examples or objectives in the continuous setting \citep{daulton2021multi, maddox2021optimizing}.
We make use of NEHVI and MTGPs, including an efficient MTGP posterior sampling approach developed in \citet{maddox2021bayesian}.

\section{Latent-Space Multi-Objective BayesOpt}
\label{sec:lambo_method_section}

We now describe the key ideas behind LaMBO,
summarized in Figure \ref{fig:optimizer_diagram} and Algorithm \ref{alg:lambo}.

\subsection{Architecture Overview}
We use a non-autoregressive denoising autoencoder to map discrete sequences to and from sequences of continuous token-level embeddings, with a multi-task GP head operating on pooled sequence-level embeddings.
Unlike previous work combining GPs with deep generative models, we do not require a pretrained autoencoder, nor do we require the surrogate to be completely reinitialized after receiving new data.
Furthermore, we demonstrate that both stochastic variational and exact GP inference can be used, alleviating concerns regarding computational scalability or applicability to noisy objectives with non-Gaussian likelihoods.
We implement our models in BoTorch \citep{balandat_botorch_2020}, and GPyTorch \citep{gardner2018gpytorch}. 
See Appendix \ref{sec:implementation_details} for more details.

\textbf{Shared Encoder:} we use a non-autoregressive bidirectional encoder $g(\vec x, \theta_{\mathrm{enc}}) = [\vec z_1, \dots, \vec z_t] = Z$, where $\vec z_i \in \mathbb{R}^d$ are latent token-level embeddings. 
In particular, our encoder is composed of 1D CNN layers, using standard vocabulary embeddings and sinusoidal position embeddings, padding token masking, skip-connections, and layernorm.
A key advantage of using a DAE rather than a VAE is that projects like ChemBERTa, TAPE and ESM have already openly released large DAE models trained on large sequence corpora \citep{chithrananda2020chemberta, rao2019evaluating, rives2021biological}.
As a result, encoders from these models could be used as drop-in replacements for our small CNN encoder.

\textbf{Discriminative Head:} this head takes an encoded sequence $Z$ and outputs a scalar value indicating the utility of selecting that sequence as a query point.
We pass $Z$ to a discriminative encoder $w$ to obtain transformed embeddings $Z'$, then pool $Z'$ into low-dimensional sequence-level features.
The discriminative encoder is smaller than the shared encoder, for example a single residual CNN block.
The pooling operation $(|\mathcal{I}(\vec x)|)^{-1}\sum_{i \in \mathcal{I}(\vec x)} \vec z'_i$ averages token-level embeddings over a restricted index set $\mathcal{I}(\vec x)$ which excludes positions corresponding to padding tokens.
We define a multi-task GP kernel by combining a $5/2$ Mat\'ern kernel evaluated on these sequence features with an intrinsic model of coregionalization (ICM) kernel over $f_i$ \citep{goovaerts1997geostatistics,alvarez2011kernels,rasmussen2006gaussian}.
The resulting GP outputs a posterior predictive distribution $p(f | \mathcal{D})$, which is passed as input to the acquisition function.
We use the noisy expected hypervolume improvement (NEHVI) acquisition from \citet{daulton2021parallel} since some objectives (particularly those involving biological data) are inherently noisy.

\textbf{Generative Decoder Head:} this head ($h$) maps a sequence of token-level embeddings to a predictive distribution over $\mathcal{X}$ and can either be a simple masked language model (MLM) head \citep{devlin2018bert} or a full seq2seq latent non-autoregressive neural machine translation (LANMT) decoder \citep{shu2020latent}.
In this work we make use of both.
An MLM head is simpler and easier to control than an LANMT decoder, which allows for careful comparisons between LaMBO and discrete genetic optimizers.
Despite their complexity, LANMT decoders accommodate insertions and deletions more gracefully than MLM heads by means of a length prediction head and length transform operation, allowing the same latent representation to be decoded to sequences of varying length.
In either case, the decoder uses the same layer types as the shared encoder $g$, and takes both $Z$ and $Z'$ as input.

\subsection{Initializing the Latent Embeddings}
\label{subsec:inner_loop_init}

At each outer-loop iteration $i$, there are many choices of $\vec x$ we are confident will \textit{not} improve our current solution $\mathcal{P}_i$, corresponding to large flat regions of the inner-loop loss surface.
If the inner-loop is not initialized carefully, it is very likely the initial solution will fall in one of these flat regions, severely hampering gradient-based optimization.
If $\vec x$ was continuous, we could use initialization heuristics from previous work to avoid this pitfall \citep{balandat_botorch_2020, daulton2021parallel}.
Instead we now show how the the same corruption process used to train and sample from DAEs can be used as a robust initialization procedure for discrete $\vec x$ by initializing the inner loop solution in latent-space ($Z_0$) with corrupted, encoded variants of the current outer-loop solution $\mathcal{P}_i$.

\begin{figure*}
    \centering
    \includegraphics[width=0.6\textwidth]{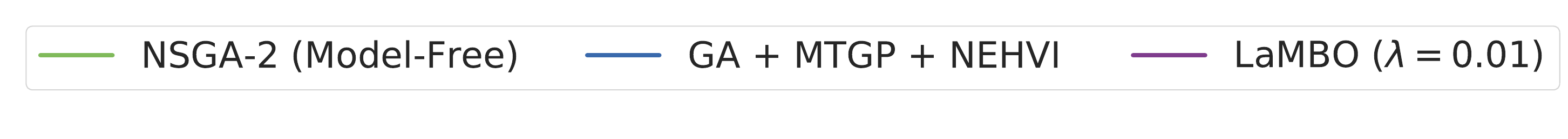}
    \vspace{-0.1in}
    \\
    \begin{tabular}{cccc}
        \includegraphics[width=0.22\textwidth]{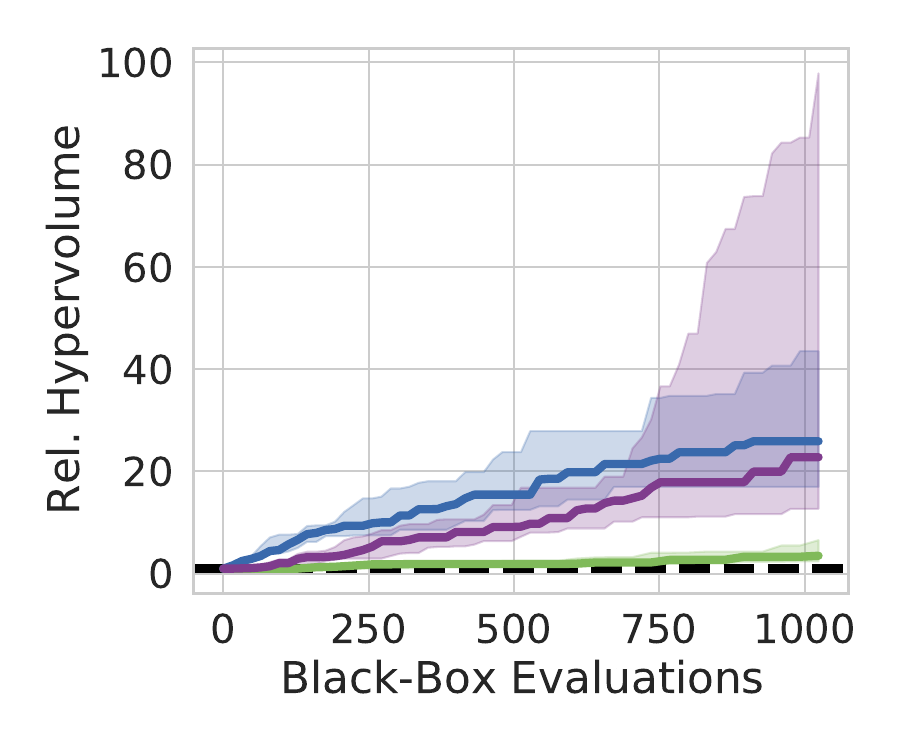}
        &
        \includegraphics[width=0.22\textwidth]{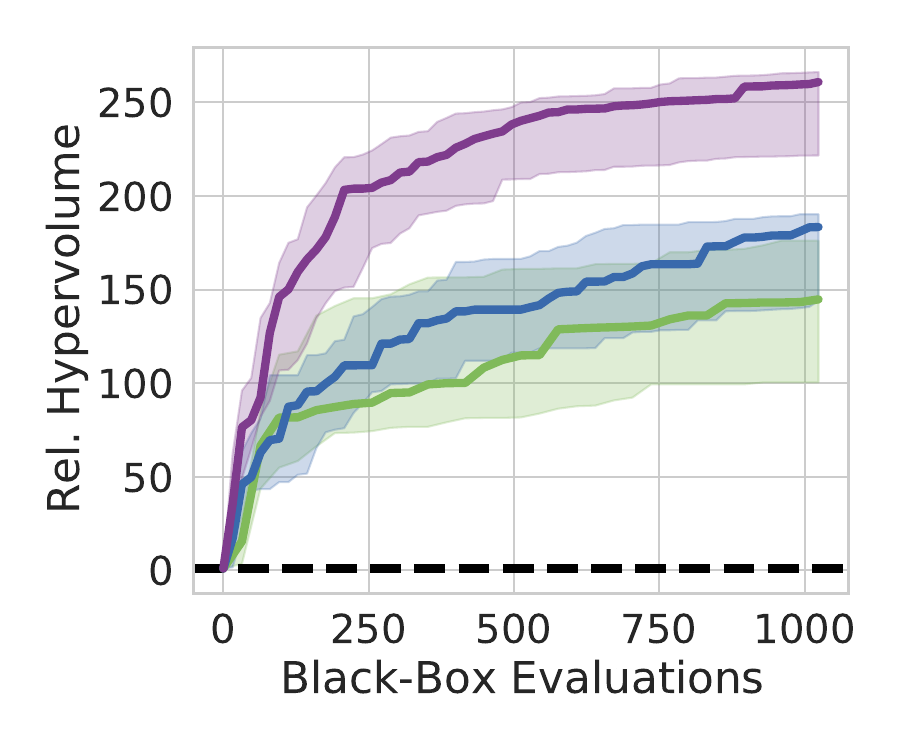}
        & 
        \includegraphics[width=0.22\textwidth]{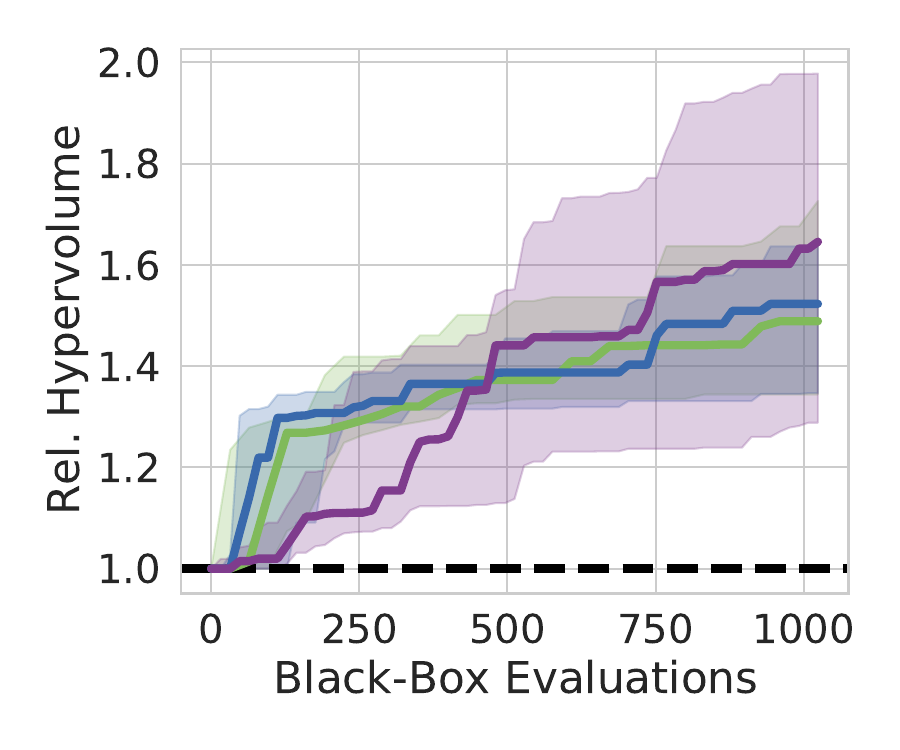}
        & 
        \includegraphics[width=0.22\textwidth]{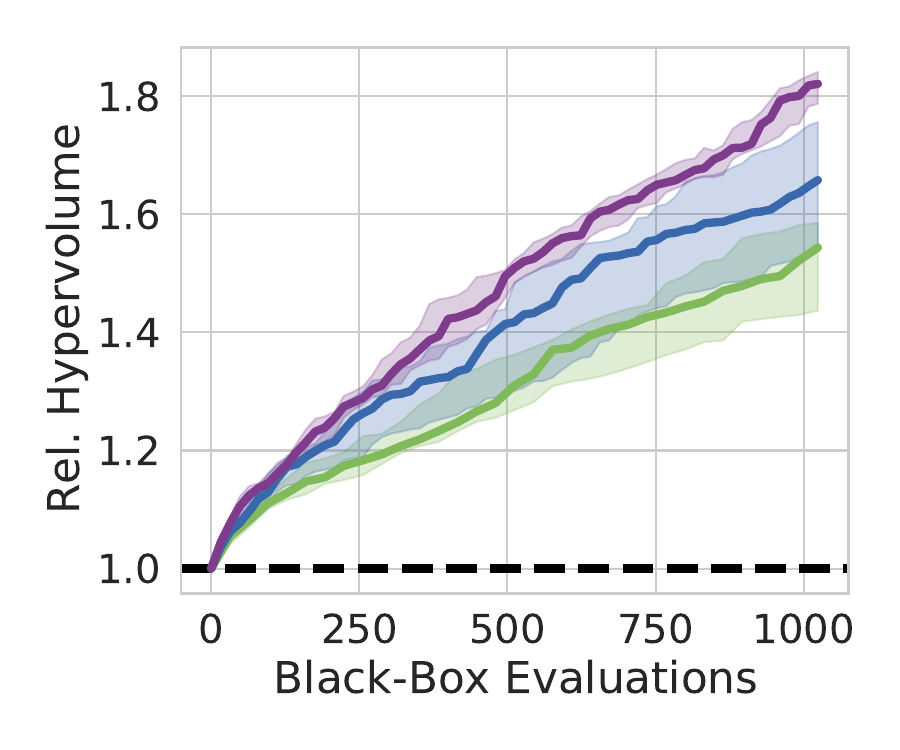} \\
        {\small \textbf{(a)} \textbf{Bigrams}} 
        & 
        {\small \textbf{(b)} \textbf{logP + QED}}
        & 
        {\small \textbf{(c)} \textbf{DRD3 Docking + SA}}
        &
        {\small \textbf{(d)} \textbf{Stability + SASA}}
    \end{tabular}
    \caption{
    On all four tasks (described in Section \ref{subsec:eval_procedures}), LaMBO outperforms genetic algorithm baselines, specifically NSGA-2 and a model-based genetic optimizer with the same surrogate architecture (GA + MTGP + NEHVI). Performance is quantified by the hypervolume bounded by the optimized Pareto frontier.
    The midpoint, lower, and upper bounds of each curve depict the 50\%, 20\%, and 80\% quantiles, estimated from 10 trials. See Section \ref{subsec:genetic_comparison} for more discussion. 
    }
    \label{fig:hypervol_rel_comparison}
\end{figure*}

\textbf{Selecting base sequences:} we begin with a set of seed sequences $\SeqPool$ and select a subset, $\SeqBase \subset \SeqPool$.
After each optimization round, the latest query sequences are added to $\SeqPool$ and can serve as future base sequences.
Choosing $\SeqBase$ well is critical for fast convergence.
If too many low quality sequences are added, too much computation is spent optimizing them.
Conversely, only selecting the current Pareto sequences could hinder exploration of sequences with potential for improvement.
The interaction between the online queries and the decoder head must also be considered, since we want to prevent the generative samples from collapsing to a couple sequences.

We populate $\SeqBase$ first with the current Pareto sequences $\mathcal{P}_i$, then with random sequences (without replacement) that were on the Pareto frontier in previous optimization rounds ($\mathcal{P}_{<i}$), and finally fill any remaining space in the base set with random sequences from the entire optimization history. 
In practice, we took the size of the base set to be the same as the query batch size $b$.
We perform multiple restarts during the inner loop, and each restart samples $b$ sequences from $\SeqBase$, with replacement.
We do not sample any base sequences uniformly at random, but according to a weighted distribution $\Delta(\SeqPool)$ to ensure that high-scoring sequences are more likely to be optimized than low-scoring ones.

Let $r_i(\vec x_j, X)$ be the rank of $\vec x_j \in X$ w.r.t. the 0-indexed dense ranking of $X$ according to $f_i$, and let $r_{\mathrm{max}}(\vec x_j, X) = \max_i r_i(\vec x_j, X)$.
The sampling weight $w_j$ of $\vec x_j \in \SeqPool$ is 
\begin{align}
    w_j &= s_j(-\log(1 + \vec r) / \tau), \\
    \vec r &= [r_{\mathrm{max}}(\vec x_1, X), \dots, r_{\mathrm{max}}(\vec x_p, X)], \nonumber
\end{align}
where $s_j(\vec v) = \exp(v_j) / \sum_i \exp(v_i)$ is the softmax function and $\tau \in (0, +\infty)$ is the softmax temperature.
In other words, we choose the least favorable ranking across all objectives for each $\vec x$ to determine its weight.
This type of weighting is similar in spirit to a procedure used by \citet{tripp2020sample} to bias a VAE loss in favor of high-scoring sequences.

\begin{algorithm}[t]
\SetAlgoLined
\textbf{Inputs:} acquisition $a$, corruption $c$, shared encoder $g$, discriminative encoder $w$, decoder $h$, base sequences $\SeqBase$, and batch size $b$. \\
$v^* \leftarrow +\infty$ \\
$Z_0 = \{g\circ c(\vec x_0), \dots, g\circ c(\vec x_b)\}, \; \vec x_m \in \SeqBase$ \\
\For{$j = 0, \dots, j_{\max}$}{
    \textcolor{blue}{$Z_j' = w(Z_j)$} \\
    $Z_{j + 1} = Z_j - \eta \nabla_Z \left [ a(Z_j') \textcolor{blue}{- \lambda \mathbb{H}\left [ h(Z_j, Z_j') \right ]} \right ]$ \\
    \textcolor{blue}{$\SeqCand \leftarrow \{\vec x'_1, \dots, \vec x'_b \} \sim h(Z_j, Z_j')$} \\
    \textcolor{blue}{$Z_{\mathrm{cand}} \leftarrow [g(\vec x'_1), \dots, g(\vec x'_b)]^\top$} \\
    \textcolor{blue}{$v_j = a(w(Z_{\mathrm{cand}}))$} \\
    \textcolor{blue}{\If{$v_j < v^*$}{
        $v^*, \mathcal{X}^* \leftarrow v_j, \SeqCand$
    }}
}
\Return $\mathcal{X}^*$
\caption{The LaMBO inner loop. For clarity, steps where LaMBO differs from LSBO are shown in \textcolor{blue}{blue}.}
\label{alg:lambo}
\end{algorithm}

\textbf{Selecting base sequence positions:} after obtaining $\SeqBase$ we apply a corruption function $c$ to each element before passing them as input to the encoder, similar to the procedure proposed in \citet{gligorijevic2021function}.
The corruption function first selects positions in each sequence to modify, then selects a modification (substitution, insertion, deletion) at those positions.
We uniformly sample positions not occupied by utility tokens, such as padding tokens.

\textbf{Selecting corruption operations:}
once sequence positions have been selected, the corruption function chooses corruption operations to apply at those positions. 
For LaMBO, the corruption procedure differs depending on whether the decoder is an MLM head or a LANMT head.
If the decoder is an MLM head then all operations are substitution operations, and the replacement tokens are all masking tokens. 
If the decoder is an LANMT head then the operation type is chosen randomly and replacement tokens are sampled uniformly from $\mathcal{V}$.
Note we use a similar corruption procedure to train the decoder head.

\subsection{Sequence Optimization}
\label{subsec:lambo_components}

At a high level, we solve the inner-loop by treating the output of the decoder head $h$ as a proposal distribution.
We iteratively refine the proposal distribution by following a regularized acquisition gradient in latent space, drawing and scoring batches of sequences along the way (Algorithm \ref{alg:lambo}).

More precisely, once we have selected and corrupted the base sequences, we pass them through the encoder to produce latent embeddings $Z_0$ that serve as the initial solution for the inner-loop optimization problem.
Then for each optimization step $0 \leq j < j_{\mathrm{max}}$, we take $Z_{j+1} = Z_j - \eta \nabla_Z [\ell_{\mathrm{query}}(Z_j)]$, where $Z'_j = w(Z_j)$ (i.e. the output of the discriminative encoder $w$),
\begin{align}
    \ell_{\mathrm{query}}(Z_j) &= a(Z_j') + \lambda \mathbb{H}[h(Z_j, Z_j')], \label{eq:reg_query_loss}
\end{align}
$\mathbb{H}$ is the Shannon entropy, and $\eta$, $\lambda$  are hyperparameters controlling the step-size and regularization strength.
We sample $\mathcal{X}_{\mathrm{batch}} = \{\vec x_0', \dots, \vec x_b'\} \sim h(Z_j, Z_j')$, and score $\mathcal{X}_{\mathrm{batch}}$ by passing it \textit{uncorrupted} through the shared encoder and discriminative head.
If we see that $\SeqBatch$ has the best acquisition value so far, we store it and continue optimizing.
Note that this procedure produces a different result than simply optimizing $Z$ and decoding once at the end.
Recall that the decoder is stochastic, and the ultimate goal of the inner loop is to produce \textit{sequences} with high acquisition value, not just high-scoring latent embeddings. 

We observed that following the unregularized acquisition gradient caused the decoder entropy to quickly increase, resulting in very uniform proposal distributions.
When the proposal distributions are uniform, LaMBO essentially performs a variant of random search.
However, because $Z_0$ is produced in the same way the decoder head is trained (i.e. the same corruption process), we expect that it will be close to other latent embeddings seen during training, and the decoder entropy will be relatively low.
These observations motivated us to include the proposal entropy penalty in Eq. \eqref{eq:reg_query_loss}, to encourage $Z_j$ to stay in a region of latent space where the decoder has non-uniform predictive distributions.

We also considered choosing a small step-size $\eta$ to implicitly confine $Z_j$ to a small region around $Z_0$, but we found it difficult to choose $\eta$ both large enough to improve the $\ell_{\mathrm{query}}$ and small enough to prevent uniform proposals. 
The proposal entropy penalty also has the added benefit of smoothing the query loss surface when using acquisitions like NEHVI, which helps make the inner loop less dependent on a good initialization.
Adding a regularization term for improved control of the acquisition function has been previously studied in the continuous setting by \citet{shahriari2016unbounded}.

\begin{figure*}
    \centering
    \includegraphics[width=0.8\textwidth,trim=0cm 0.6in 0cm 0.25in, clip]{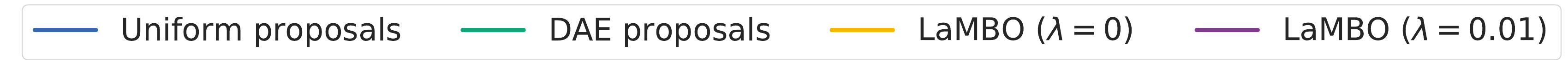}
    \\
    \begin{tabular}{cccc}
        \includegraphics[width=0.22\textwidth]{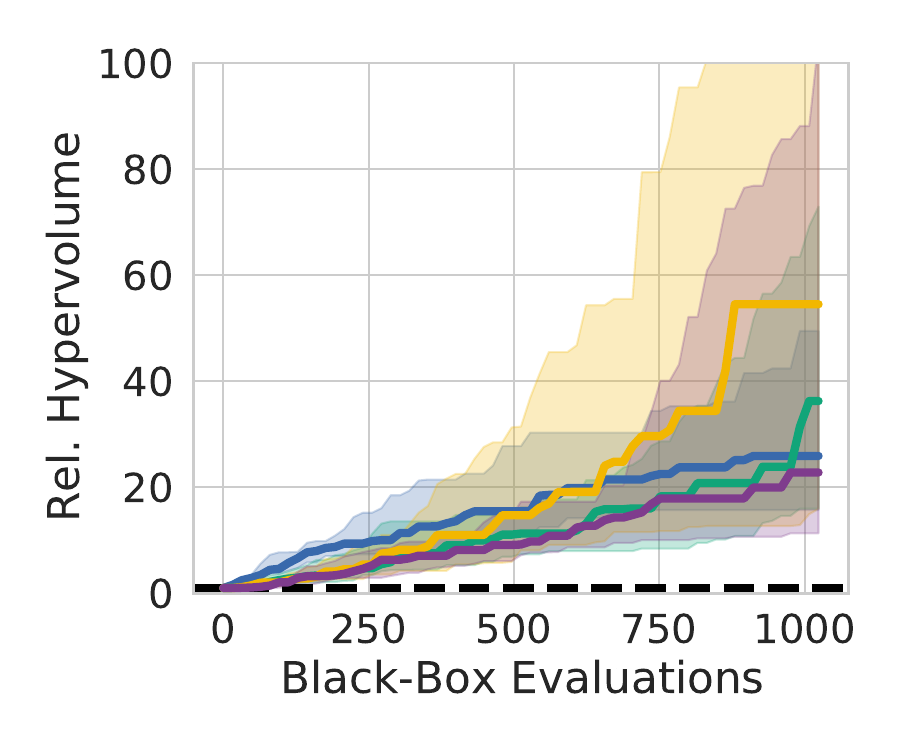}
        &
        \includegraphics[width=0.22\textwidth]{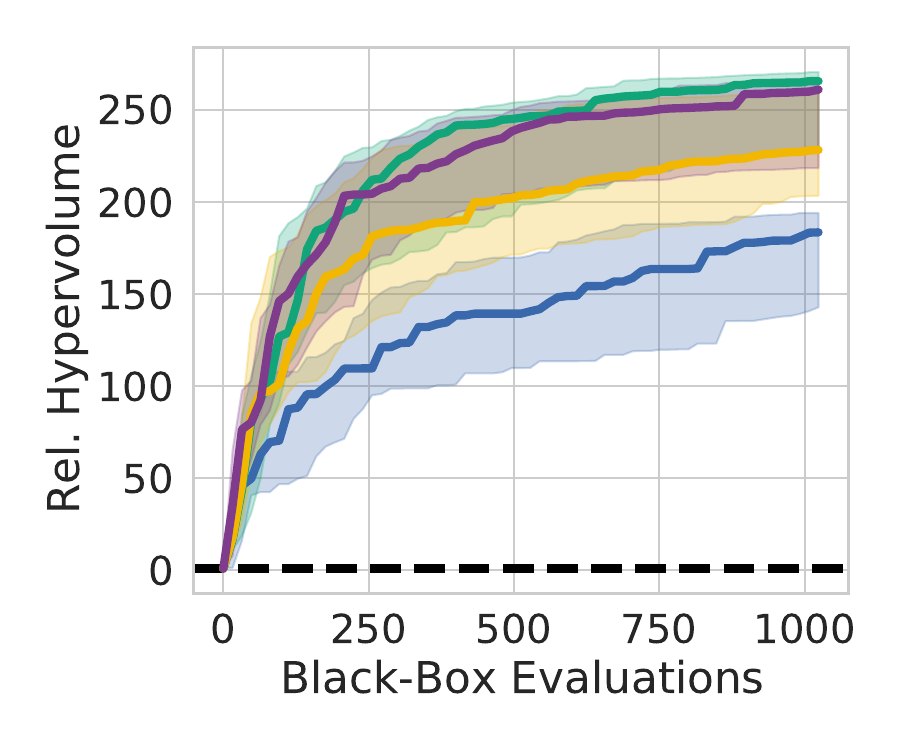}
        &
        \includegraphics[width=0.22\textwidth]{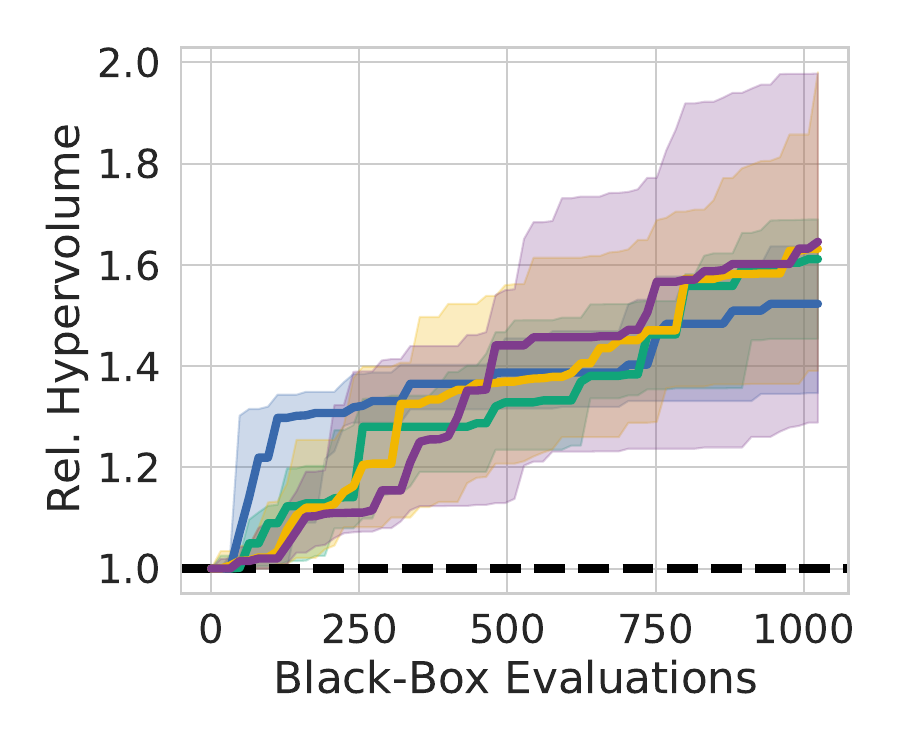}
        & 
        \includegraphics[width=0.22\textwidth]{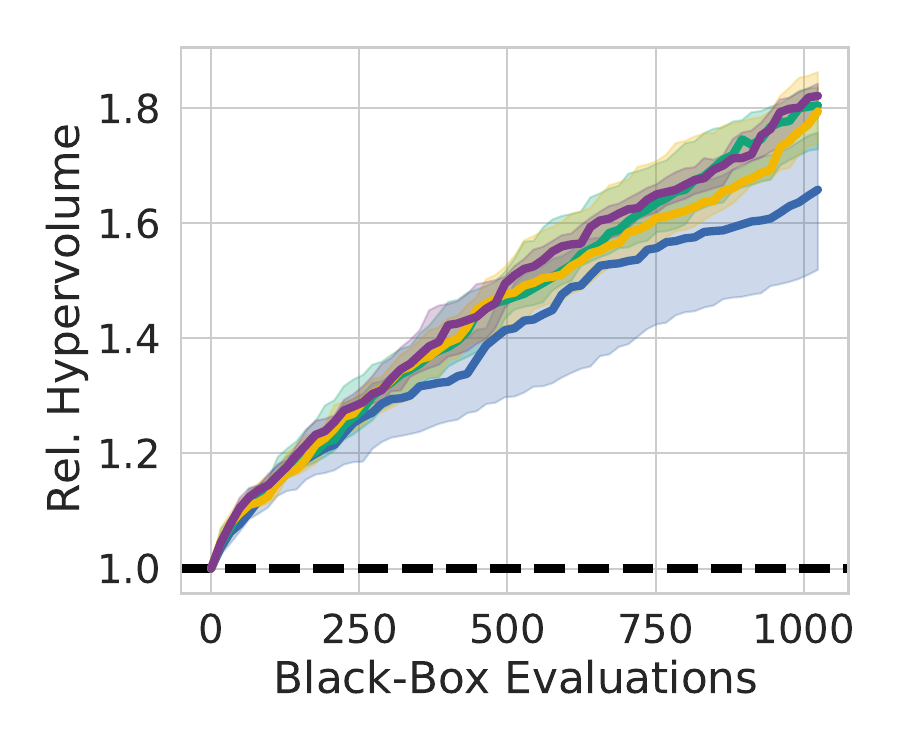} \\
        {\small \textbf{(a)} \textbf{Bigrams}} 
        & 
        {\small \textbf{(b)} \textbf{logP + QED}}
        & 
        {\small \textbf{(c)} \textbf{DRD3 Docking + SA}}
        &
        {\small \textbf{(d)} \textbf{Stability + SASA}}
    \end{tabular}
    \caption{An ablation of LaMBO's main components.
    Starting from our model-based genetic algorithm baseline (uniform proposals),
    we cumulatively add the elements described in Section \ref{subsec:lambo_components}: (1) DAE-generated proposals, (2) DAE proposal optimization following $\nabla_Z[\ell_{\mathrm{query}}]$ with $\lambda = 0.$ (see Eq. \eqref{eq:reg_query_loss}), and (3) DAE proposal optimization with $\lambda = 0.01$.
    DAE proposals improve performance on all tasks.
    Proposal optimization is most helpful on \textbf{Bigrams} where the starting sequence distribution is very unlikely to produce high-scoring queries.
    The entropy penalty is detrimental when working with random sequences \textbf{(a)}, but is helpful when working with biological sequences \textbf{(b-d)}.
    The midpoint, lower, and upper bounds of each curve depict the 50\%, 20\%, and 80\% quantiles, estimated from 10 trials.
    }
    \label{fig:lambo_ablation}
\end{figure*}

\section{Empirical Evaluation}
\label{sec:empirical_evaluation}

We now evaluate LaMBO on a suite of small-molecule and large-molecule sequence design tasks.
In Section \ref{subsec:eval_procedures} describe our suite of \textit{in silico} tasks, including a new multi-objective large-molecule task which in which we maximize the folding stability and SASA of RFPs.
See Appendix \ref{subsec:ssk_offline_regression} for an experiment comparing SSK GPs and DKL GPs, and Appendix \ref{subsec:lsbo_comparison} for an experiment comparing LSBO and LaMBO.
In Section \ref{subsec:genetic_comparison} we show that LaMBO outperforms strong genetic algorithm baselines in a carefully controlled comparison, followed by two investigative experiments in Section \ref{subsec:lambo_ablation} which give insight into the design choices behind LaMBO.
Finally in Section \ref{subsec:analyzing_designs} we analyze molecules designed by LaMBO, including \textit{in vitro} wet lab results showing the discovery of improved RFP variants.

\subsection{Evaluation procedure}
\label{subsec:eval_procedures}

We consider the following \textit{in silico} evaluation tasks, with full descriptions in the appendix:
\begin{itemize}
    \item \textbf{Bigrams}: optimize short strings (around 32 tokens) uniformly sampled from $\mathcal{X}$ to maximize the counts of three predetermined bigrams (Appendix \ref{subsec:bigram_task_details}).
    \item \textbf{logP + QED:} optimize SELFIES-encoded small molecules (50-100 tokens) w.r.t. logP and QED (Appendix \ref{subsec:chem_task_details}).
    \item \textbf{DRD3 docking + SA:} optimize SELFIES-encoded small molecules (50-100 tokens) w.r.t. DRD3 docking and synthetic accessibility (SA) (Appendix \ref{subsec:docking_task_details}).
    \item \textbf{Stability + SASA:} optimize large-molecule RFPs (around 200 tokens) w.r.t. folding stability and SASA.
    Both objectives require the 3D protein structure to compute, however we treat the folding simulator as part of the black-box objective (Appendix \ref{subsec:proxy_rfp_details}). 
\end{itemize}

Unless otherwise noted, each task begins with 512 examples in its start pool, and collects a total of 1024 online queries in batches of 16. 
No additional pretraining data is used.
Each method uses the same architecture and hyperparameters for all tasks (Appendix \ref{sec:implementation_details}).
We evaluate all methods by comparing the relative improvement of the hypervolume bounded by the Pareto front after $x$ black-box function calls compared to the starting hypervolume.

\begin{figure*}
    \centering
    \begin{tabular}{ccc}
        \includegraphics[width=0.28\textwidth]{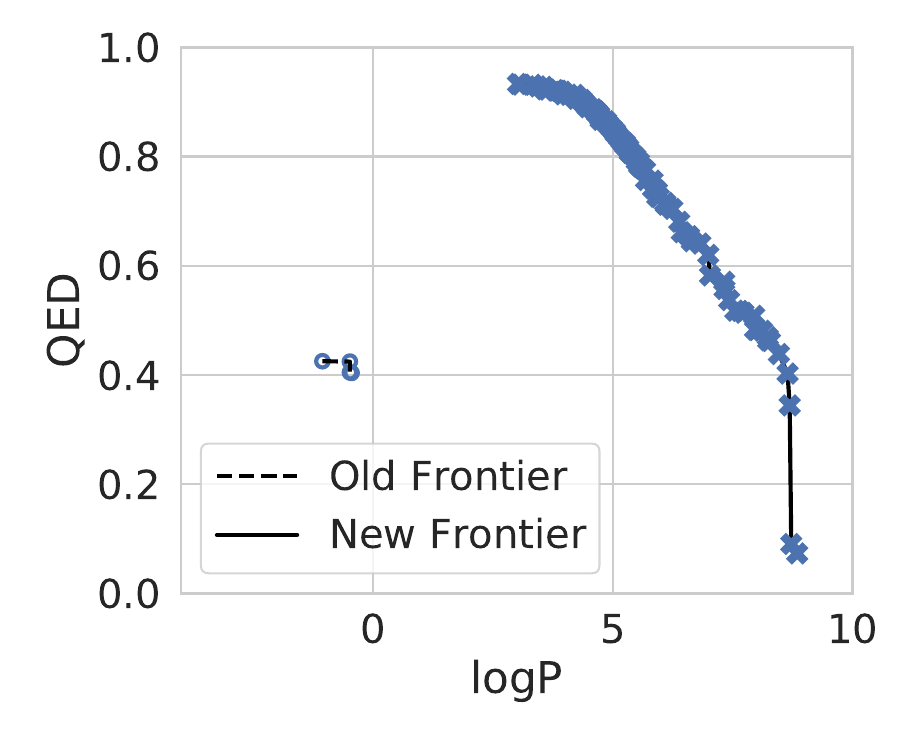}
         &
        \includegraphics[width=0.28\textwidth]{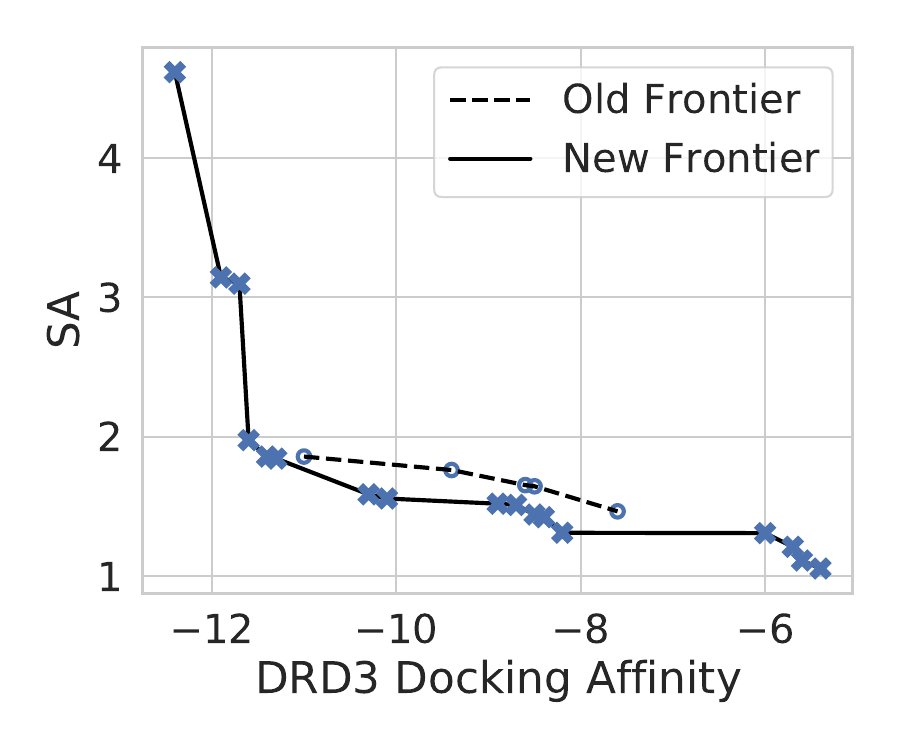}
         & 
        \includegraphics[width=0.28\textwidth]{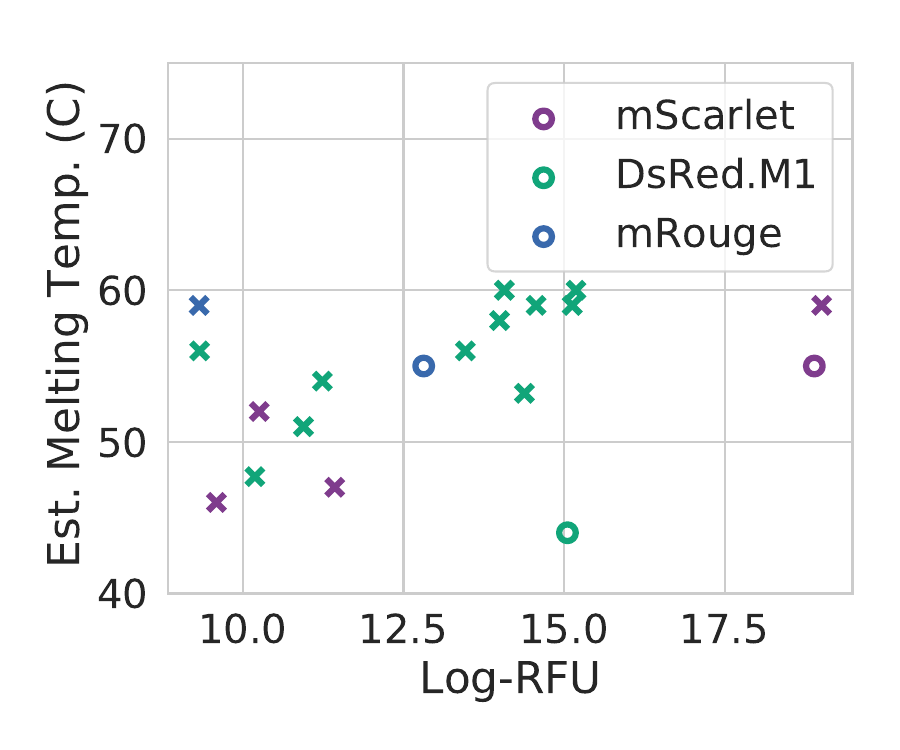}
         \\
         \textbf{(a) logP + QED} &
         \textbf{(b) DRD3 docking + SA} &
         \textbf{(c) RFP Wet Lab Results}
    \end{tabular}
    \caption{
    \sds{\@BigHat, please review}
    Here we show the old and new Pareto fronts in objective space discovered by LaMBO in Section \ref{subsec:genetic_comparison}. For \textbf{(a)} higher is better for both objectives, and for \textbf{(b)} lower is better.
    For all tasks every point in the old frontier is strictly dominated by at least one point in the new frontier, and solutions are evenly spread across the new frontiers.
    In \textbf{(c)} we provide wet lab measurements of brightness ($x$ axis) and thermostability ($y$ axis) for proteins optimized for \textbf{Stability + SASA}. Higher is better for both objectives. We discovered new, non-dominated protein variants for three of the five ancestor proteins in our evaluation set.
    }
    \label{fig:docking_pareto_frontier_viz}
\end{figure*}

\subsection{Comparing to Multi-Objective Genetic Optimizers}
\label{subsec:genetic_comparison}

In Figure \ref{fig:hypervol_rel_comparison} we compare LaMBO with a MLM decoder head against two genetic algorithm (GA) baselines. 
For this experiment we set the entropy penalty at $\lambda = 0.01$.
The simplest baseline is NSGA-2, a robust model-free multi-objective GA, which effectively simply randomly mutates solutions along the Pareto frontier. 
The other baseline is a model-based GA which also randomly mutates solutions but also screens new queries with a discriminative model, for which we use the same architecture and acquisition function (NEHVI) as LaMBO.
Both GA baselines use a uniform mutation proposal distribution.
In this experiment all optimizers are only allowed to change a single token per optimization round, and each optimizer selects base sequences and token positions in the same way. 
The model-based GA differs from LaMBO primarily in two respects: (1) the encoder is trained only through the supervised loss, and (2) the proposal distribution is uniform rather than generated by a DAE.
In fact LaMBO can be viewed as a generalization of the model-based GA, where the proposal distribution is learned and optimized, rather than fixed \textit{a priori}.
The effect of (2) is most strongly seen in \textbf{logP + QED}, since the task requires very little exploration and the SELFIES vocabulary for small molecules is significantly larger than the amino acid vocabulary for proteins.
LaMBO performs well on all four tasks, particularly those involving natural sequences \textbf{(b-d)}.
In contrast the starting distribution over sequences in the \textbf{Bigrams} task has the highest possible entropy, so LaMBO learns a good sampling distribution more slowly.

\subsection{Analyzing LaMBO}
\label{subsec:lambo_ablation}

\begin{figure}[t!]
    \centering
    \includegraphics[width=0.28\textwidth]{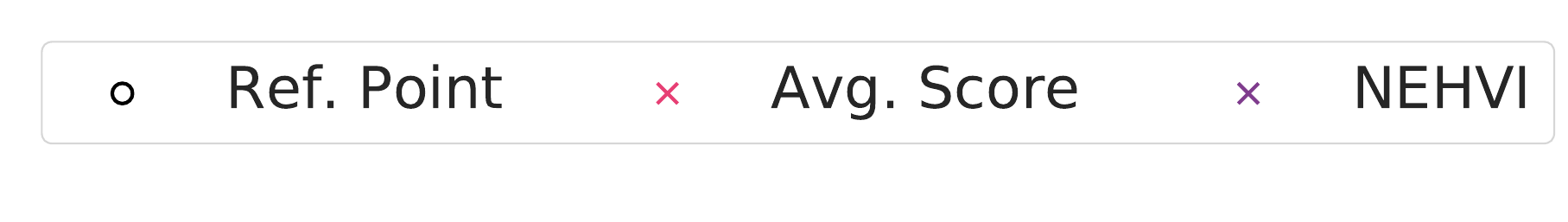}
    \includegraphics[height=0.2\textwidth]{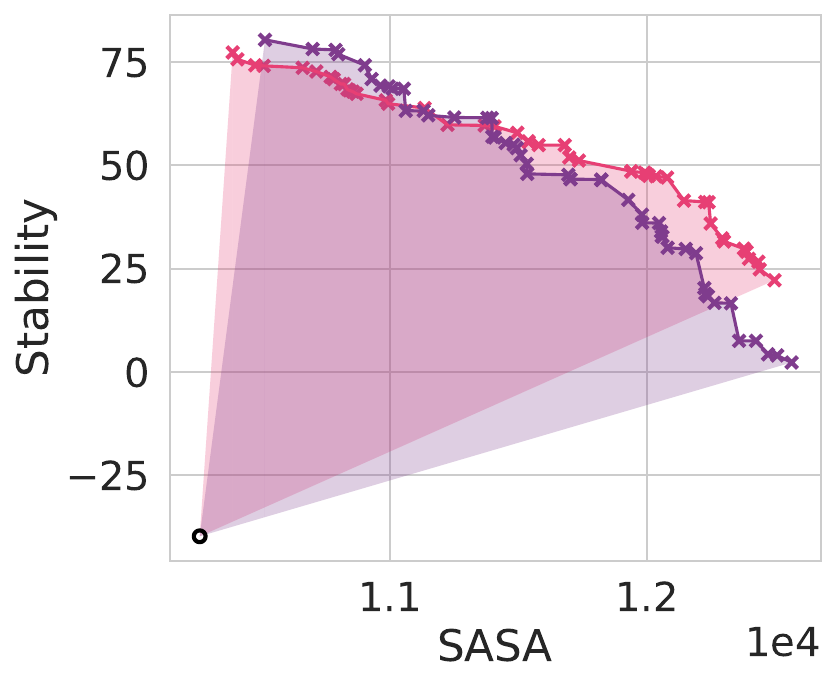}
    \caption{
    Pareto fronts of LaMBO and a variant that optimizes the expected average objective value on the RFP task.
    NEHVI incentivizes more exploration of the frontier extremities, resulting in solutions representing more diverse functional tradeoffs and slightly higher hypervolume.
}
    \label{fig:acquisition_sensitivity}
\end{figure}

Having demonstrated that LaMBO compares favorably to genetic optimizers, we now examine the different components of LaMBO and how 
each contributes to performance. 
We first disentangle the effect of replacing a uniform proposal distribution with a DAE-generated one, and the effect of optimizing the proposal distribution by gradient descent on $\ell_{\mathrm{query}}$.
In Figure \ref{fig:lambo_ablation} we interpolate between the model-based GA baseline in Figure \ref{fig:hypervol_rel_comparison} and LaMBO by cumulatively adding DAE-generated proposals, proposal optimization, and the proposal entropy penalty.
We find that unoptimized DAE proposals with NEHVI screening is a strong baseline on all tasks.
Proposal optimization is the most useful in \textbf{Bigrams}, where the true non-dominated solutions $\mathcal{P}^*$ lie far outside the starting sequence distribution, requiring more exploration.
For the same reason the entropy penalty is somewhat detrimental specifically in \textbf{Bigrams}, since it keeps new queries near those previously seen.
See Figure \ref{fig:lambo_ablation_extras} in Appendix \ref{sec:additional_results} for more discussion.

In Figure \ref{fig:acquisition_sensitivity}, we evaluate the sensitivity of LaMBO to the choice of acquisition function on the RFP task.
We compare the final Pareto frontier obtained with a simple multi-objective scalarization (averaging normalized scores) to the frontier obtained with NEHVI.
Score averaging focuses optimization on solutions with similar tradeoffs, pushing the interior of the frontier out quickly.
This behavior leads to less exploration than NEHVI and thus a slightly lower hypervolume of $1.51$ as compared to NEHVI's hypervolume of $1.57.$
Our findings corroborate similar results in previous work comparing scalarization and hypervolume-based acquisitions in different problem settings \citep{emmerich2005single,emmerich2011hypervolume,daulton_differentiable_2020}.

\begin{figure*}
    \centering
    \begin{tabular}{|ccccc|}
        \hline
        & $\leftarrow$ \textbf{better logP} & & \textbf{better QED} $\rightarrow$ & \\
        \hline
        \includegraphics[width=0.17\textwidth]{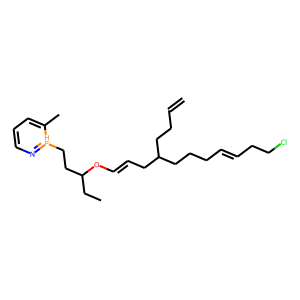}
        &
        \includegraphics[width=0.17\textwidth]{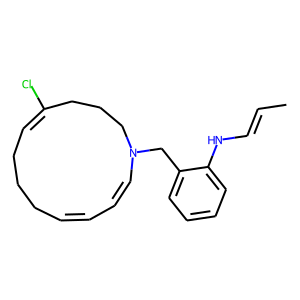}
        &
        \includegraphics[width=0.17\textwidth]{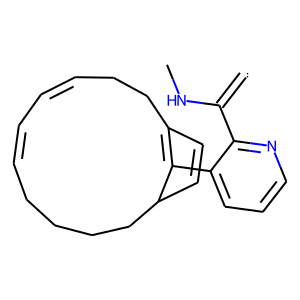}
        &
        \includegraphics[width=0.17\textwidth]{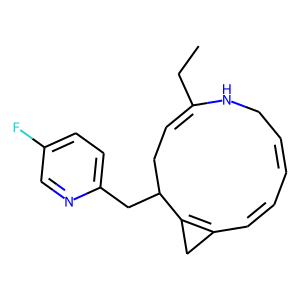}
        &
        \includegraphics[width=0.17\textwidth]{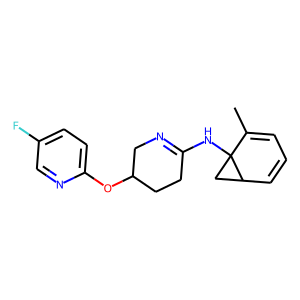}
        \\
        \hline
        \textbf{7(a)-1:} (8.84, 0.07) & \textbf{7(a)-2:} (6.59, 0.65) & \textbf{7(a)-3:} (5.35, 0.81) & \textbf{7(a)-4:} (4.48, 0.89) & \textbf{7(a)-5:} (3.02, 0.93) \\
        \hline
        \hline
        & $\leftarrow$ \textbf{better docking} &  & \textbf{better SA} $\rightarrow$ & \\
        \hline
        \includegraphics[width=0.17\textwidth]{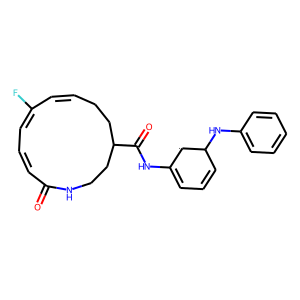}
        &
        \includegraphics[width=0.17\textwidth]{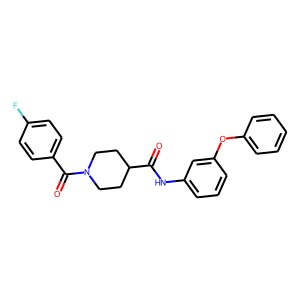}
        &
        \includegraphics[width=0.17\textwidth]{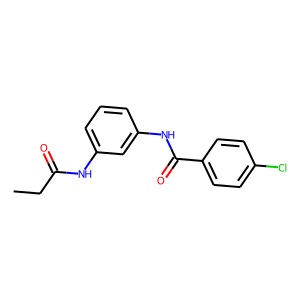}
        &
        \includegraphics[width=0.17\textwidth]{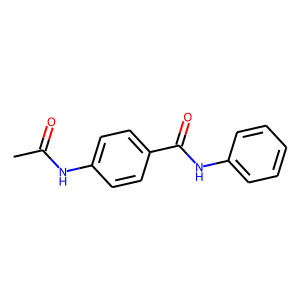}
        &
        \includegraphics[width=0.17\textwidth]{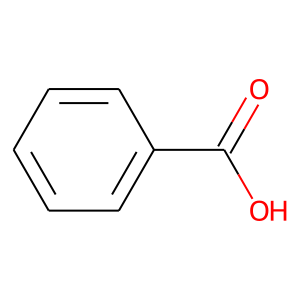}
        \\
        \hline
        \textbf{7(b)-1:} (-12.4, 4.61) & \textbf{7(b)-2:} (-11.4, 1.86) & \textbf{7(b)-3:} (-8.9, 1.52) & \textbf{7(b)-4:} (-8.2, 1.31) & \textbf{7(b)-5:} (-5.4, 1.05) \\
        \hline
    \end{tabular}
    \caption{\sds{\@BigHat, please review} Example molecules at varying points on the new Pareto frontier discovered by LaMBO in Section \ref{subsec:genetic_comparison} for \textbf{logP + QED} (top) and \textbf{DRD3 docking + SA} (bottom). Under each molecule we show the objective values as $\mathrm{index}: (y_1, y_2)$. From left to right $y_1$ worsens and $y_2$ improves.
    The Pareto frontier is a rich, low-dimensional space that can be analyzed much more easily than the whole search space.
    }
    \label{fig:drd3-sa_mol-viz}
\end{figure*}

\subsection{Analyzing Sequence Designs}
\label{subsec:analyzing_designs}

Though metrics like hypervolume are useful for conducting baseline comparisons and ablation studies, such metrics do not give much insight into more qualitative aspects of the sequences we are designing.
In Figure \ref{fig:docking_pareto_frontier_viz}(a-b) we show the Pareto fronts found by LaMBO for our two small-molecule tasks, as we did in Figure \ref{fig:intro_pareto_frontier} for the large-molecule task.
For every task every sequence on the old frontier is strictly dominated by at least one sequence on the new frontier.

So far we have shown that LaMBO can optimize \textit{in silico} objectives effectively, and argued in Appendix \ref{subsec:proxy_rfp_details} that the \textbf{Stability + SASA} objectives are likely correlated with properties of RFPs we actually wish to optimize in the real world, specifically brightness and thermostability.
In Figure \ref{fig:docking_pareto_frontier_viz}(c) we evaluate \textit{in vitro} protein sequences designed by LaMBO to optimize \textbf{Stability + SASA}, reporting brightness (log relative fluorescence units, log-RFU) and thermostability (protein melting temperature).
We discovered non-dominated variants of three of the five ancestor sequences in our evaluation set, including strictly dominant variants of mScarlet and DsRed.M1, with inconclusive results for the other two ancestor sequences.
See Appendix \ref{subsec:wetlab_details} for more details about our experimental procedure and results.
Our results indicate that \textbf{Stability + SASA} are good \textit{in silico} proxy objectives for a real-world protein design task.

In Figure \ref{fig:drd3-sa_mol-viz} we visualize solutions at different points on the optimized Pareto frontiers found by LaMBO for our two small-molecule tasks.
Intriguingly, we find that although we did not specifically optimize for solution diversity in sequence space, the sequences change significantly as we move along the frontiers.
In contrast, many generative-only approaches use heuristics to explicitly encourage diversity in sequence space and prevent solution collapse.

Some of the proposed molecules shown have features such as large macrocycles in 7(a)-2 and 7(b)-1, or three-carbon rings in 7(a)-4 and 7(a)-5, that are likely not synthetically accessible \citep{gao2021amortized}. 
However, we note that \textbf{logP + QED} does not explicitly incentivize accessibility, and remarkably one can deduce that molecules with large macrocycles are difficult to synthesize even with little knowledge of chemistry by simply comparing molecules 7(b)-1 and 7(b)-2.
Our results highlight the multi-objective nature of drug design, the need for careful objective selection (i.e. target discovery), and the human-interpretable insights that can be gained by studying the differences between non-dominated solutions found by machine learning methods.

\vspace{-2mm}
\section{Discussion}

Drug design is quickly emerging as an extremely important application of machine learning.
BayesOpt has extraordinary potential for this domain, but existing approaches struggle to extend to high-dimensional, discrete, multi-objective design tasks. 
We have shown how deep generative models can be integrated with BayesOpt to address these challenges, achieving good sample efficiency and solution quality across a range of design tasks.
Moreover, we introduced a new large-molecule task, which provides a challenging and realistic benchmark with which to evaluate new methods for biological sequence design.
Finally we successfully optimized red fluorescent proteins \textit{in vitro}, and showed how characterizing the Pareto frontier can lead to useful, interpretable scientific insights.

In the short term, we are excited to combine LaMBO with large specialized pretrained generative models for antibodies \citep{ruffolo2021deciphering}. The selection of mutation sites in the initialization procedure in Section \ref{subsec:inner_loop_init} could also be improved beyond uniform random sampling.
In the longer term, there are also many promising developments in BayesOpt methodology that have yet to be explored for sequence design, such as non-myopic acquisition functions \citep{jiang2020efficient}, multi-fidelity acquisition functions to determine the type of experimental assay and number of replications \citep{kandasamy2017multi, wumulti2019}, rigorous treatment of design constraints \citep{eriksson2019scalable}, and the coordination of many parallel drug development campaigns by optimizing risk measures across a whole compound portfolio \citep{cakmak2020bayesian}. BayesOpt with multi-modal inputs is a particularly exciting direction, allowing scientists to combine many different sources of experimental data, including 3D structure and raw instrument output \citep{jin2021iterative}. 
BayesOpt itself can also be developed for greater resilience to model misspecification, miscalibration, and covariate shift, all of which are important in drug design tasks. Finally, our work also the highlights the need for better benchmarks to evaluate drug design methods.

While there are still many challenges to overcome, there is a path for Bayesian optimization to revolutionize drug design, profoundly improving our lives, and changing the way we approach scientific discovery.
\vspace{-2mm}
\paragraph{Acknowledgements.} We would like to thank Sait Cakmak, Andres Potapczynski and Sanyam Kapoor for helpful discussions. This research is supported by an Amazon Research Award, Facebook
Research, Google Research, Capital One, NSF CAREER IIS-2145492, NSF I-DISRE 193471, NSF CDS\&E-MSS 2134216, NIH R01DA048764-01A1, NSF IIS-1910266, and NSF 1922658 NRT-HDR.

\bibliography{references}
\bibliographystyle{icml2022}

\onecolumn
\newpage
\appendix
\appendixpage

In Appendix \ref{sec:proxy_task_details} we outline the four evaluation tasks: \textbf{Bigrams}, \textbf{logP + QED}, \textbf{DRD3 docking + SA}, and \textbf{Stability + SASA}. 
In Appendix \ref{sec:implementation_details}, we describe in detail the neural network architecture, the DKL GP implementation, the denoising autoencoder implementation, the string kernel implementation, and the hyperparameters used for our experiments. 
Finally, in Appendix \ref{sec:additional_results}, we present additional experimental results to supplement the figures in the main text.

\section{Evaluation Task Details}
\label{sec:proxy_task_details}

\subsection{Bigrams Task}
\label{subsec:bigram_task_details}

The bigrams task is a simple toy example of discrete sequence optimization. 
We draw random strings from an alphabet $\mathcal{V}$ and count the occurrence of $k$ predetermined bigrams, which we use as proxy fitness targets. 
The task is to maximize the counts of each bigram in the sequence, restricting the sequence length to 36 tokens including utility tokens ("[PAD]", "[CLS]", "[SEP]", "[UNK]", "[MASK]").
For our experiments we used the same amino acid vocabulary as our protein task and chose 3 complementary bigrams, ``AV", ``VC" and ``CA". 
The initial sequences were sampled with lengths between 32 and 36 tokens
We ensured there were an equal number of positive examples (sequences with at least one occurrence of one of the bigrams) as negative examples in the starting pool.

\subsection{logP + QED Task}
\label{subsec:chem_task_details}

The original ZINC logP optimization task, popularized in the BayesOpt community by \citet{gomez2018automatic}, is to optimize the octanol-water partition coefficient of a small molecule. Molecules with high logP values are hydrophobic and molecules with low values are hydrophilic. Hydrophobicity can be desirable for absorption and solubility, for example in pharmaceuticals. As a property that is easy to calculate, it has risen to prominence despite being undesirable on its own. Very high logP can result in molecules with limited practical application, and moreover finding molecules with high logP reduces trivially to the problem of finding long hydrocarbon chains, as these compounds are extremely hydrophobic relative to the size of the molecule.
The \textit{penalized} logP objective adds auxiliary terms measuring synthetic accessibility \citep{ertl2009estimation} and the number of cycles.
Unfortunately these terms do not fix the underlying problem, and so penalized logP is similarly vulnerable to optimization hacking, as we discuss in Section \ref{subsec:lsbo_comparison}.

Because logP in itself is a deeply flawed objective, both in its relevance to real-world drug design and its ability to be hacked by optimizers, we also consider a multi-objective optimization task that is closer in form to real design problems. Instead of solely optimizing for logP, we jointly optimize for logP and QED (Quantitative Estimate of Druglikeness), a composite metric that captures many elements of druglikeness, with bioaavailability among the most prominent \cite{bickerton2012quantifying}.
Unfortunately QED has its own limitations as an objective, since it is a simple parametric model trained on a small dataset to emulate heuristics such as Lipinski's Rule of Five \citep{lipinski1997experimental}.

The shortcomings of objectives like logP and QED appear to be well-known \citep{nigam2019augmenting, coley2020autonomous, fu2020mimosa, tripp2021fresh, maus2022local}, but superior alternatives have not yet been accepted by the research community.
For example, at the time of writing the only molecule generation benchmark in TorchDrug is maximization of QED and logP of ZINC-like molecules.\footnote{\url{https://torchdrug.ai/docs/benchmark/generation.html}}
\citet{angermueller2020population} evaluated BBO algorithms on a substantial number of \textit{in silico} sequence design tasks, however the large molecule tasks they considered were relatively simple, single-objective problems (e.g. maximization of the likelihood of a hidden Markov model).
The vacuum of rigorous \textit{in silico} evaluation tasks for large-molecule design motivated us to propose our RFP task as a new benchmark.

We construct the start pool by inverting the scores and selecting the top-$k$ non-dominating sequences (i.e. we found the $k$ \textit{most} dominated sequences in the ZINC dataset w.r.t. logP and QED). Constructing the task in this way is better than simply sampling randomly from ZINC because QED is bounded above by 1 and many ZINC sequences already score fairly close to 1. 
Starting with dominated sequences ensures that there is sufficient headroom for improvement to observe variations in optimizer behavior. We capped the max sequence length at 128 SELFIES tokens, including utility tokens. 
The SELFIES vocabulary was precomputed from the entire ZINC dataset \citep{krenn2020self}.

\subsection{DRD3 Docking + SA Task}\label{subsec:docking_task_details}

\sds{\@BigHat, please review}
We use the DRD3 docking score oracle from \citet{huang2021therapeutics}, and the following quotation is reproduced from \url{https://tdcommons.ai/benchmark/docking_group/overview}:

\say{
Docking is a theoretical evaluation of affinity between a ligand (a small molecular drug) and a target (a protein involved in the disease). As a molecule with higher affinity is more likely to have higher bioactivity, docking is widely used for virtual screening of compounds \citep{lyu2019ultra}.
}

Because optimizing solely for docking may produce molecules that are difficult to synthesize, we also optimize for synthetic accessibility (SA) \citep{ertl2009estimation}.

To construct the start pool for this task we selected 512 molecules from ZINC uniformly at random and labeled them with the objective oracles. 
We then used the same start pool and SELFIES encodings for all experimental trials and all optimization methods.

\subsection{Stability + SASA Task}
\label{subsec:proxy_rfp_details}

In this work we present a new \textit{in silico} benchmark task designed to simulate searching for improved red fluorescent protein (RFP) variants \textit{in vitro}, a problem of significant interest to biomedical researchers \citep{dance2021hunt}. We optimize red-spectrum proteins with known structures for stability (-dG or negative change in Gibbs free energy) and solvent-accessible surface area (SASA) \citep{shrake1973environment, cock2009biopython} in simulation, using the FoldX suite \citep{schymkowitz2005foldx} and BioPython to evaluate our objective function. Stability as evaluated by FoldX---particularly in the negative case---has been shown to correlate with protein function \citep{Hoie_2021}. Solvent-accessible surface area will correlate with factors that influence the brightness and photostability of the fluorescent protein: aggregation propensity due to exposed hydrophobic surface \citep{Mishra_2018} and shielding by the beta-barrel, which encapsulates the fluorophore \citep{Chudakov_2010}. Since both of these benchmark tasks are functions of the protein's three-dimensional structure, it is expected that training a model on these tasks will require the model to learn a latent representation for structure, which in turn determines function. 

We constructed the start pool in two phases. 
First we searched FPBase for all red-spectrum (defined in this context as having an emission wavelength at least 580 nm) proteins with at most 244 residues with known 3D structures, selecting the highest resolution structure if more than one was available.
If more than one chain was present in the structure, we selected the longest chain as the representative residue sequence.
Starting from these base proteins, we used NSGA-2 to collect additional labelled sequences to use in the start pool for subsequent experiments. 

Although this task is a significant step forward for \textit{in silico} evaluation of discrete sequence design, it is currently limited by the capabilities of FoldX, which can only compute structures from substitution mutations (i.e. the sequence length cannot change). 
Deep learning structure oracles such as AlphaFold \citep{jumper2021highly} or RoseTTAFold \citep{baek2021accurate} could also be used, but we found FoldX to be simpler and more amenable for rapid prototyping.

\subsection{Wet lab experimental procedure}
\label{subsec:wetlab_details}

\sds{\@BigHat, please review}
We synthesized fluorescent proteins with PUREfrex 2.1 (Cosmo Bio LTD) in 50 uL reactions from linear DNA purchased from IDT as eBlock dsDNA gene fragments. We ran reactions at 30$^{\circ}$C overnight in black, half-area microplates (Corning \#3993) with optically clear plate adhesive and measured excitation and emission through a series of sweeps (fixing the excitation wavelength and scanning emission every 1--2~nm, or vice versa). We determined peak excitation and peak emission as the wavelength that gave maximum fluorescence units. Using NanoDSF on an Uncle instrument  (Unchained Labs), we measured protein thermostability as T$_\textrm{m}$, defined as the midpoint transition temperature of the thermal melt curve.

\section{Implementation Details}
\label{sec:implementation_details}

Our models are implemented in PyTorch \citep{paszke2019pytorch}, BoTorch \citep{balandat_botorch_2020}, and GPyTorch \citep{gardner2018gpytorch}.
Our genetic optimizer baselines are implemented in PyMOO \citep{blank2020pymoo}.
Our code is publicly available at \url{https://github.com/samuelstanton/lambo}.
Hyperparameters are summarized in Appendix \ref{subsec:hypers}.

\subsection{Architecture Details}

We used the same base architecture for all experiments, relying on 1D convolutions (masking positions corresponding to padding tokens). We used standard pre-activation residual blocks with two conv layers, layernorm, and swish activations.
We used a kernel size of 5, 64 intermediate channels and 16 latent channels. 

The shared encoder and decoder each were composed of 3 of these residual blocks (for a total of 6 convolutional layers each). The shared encoder embeds input sequences with standard vocabulary and sinusoidal position embeddings.
The discriminative encoder was composed of a single residual block.

Note that transformer encoder layers could be substituted as a drop-in replacement for these convolutional residual blocks, we used small convolutional layers because they are fast to train and performed adequately in our experiments.

For multi-task GPs, we chose ICM kernels for their efficiency, particularly in sampling \citep{bonilla2007multi,maddox2021bayesian}, their flexibility and popularity.

\subsection{DKL Implementation Details}
\label{subsec:dkl_implementation_details}

Training DKL models takes care. Some best practices apply to both stochastic variational \citep{wilson2016stochastic} and exact GP inference \citep{wilson2016deep} with DKL, others are specific to the former. 

\paragraph{Applicable to exact and variational GP inference}
\begin{enumerate}
    \item \textit{Kernel hyperparameter priors matter.} Allowing the DKL GP to easily change both the inputs to the final conventional GP kernel (e.g. RBF) and the lengthscale of that kernel doesn't work well. We placed a tight Gaussian prior ($\sigma = 0.01$) around the initial lengthscale value and forced the encoder to learn features appropriate for that lengthscale. Note that this is distinct from simply fixing the kernel hyperparameters \textit{a priori}.
    \item \textit{Optimizer hyperparameters matter.} Adam is really convenient to avoid too much learning rate tuning, but it can cause unexpected issues when jointly training supervised and unsupervised heads. We almost completely disabled the running estimates of the first two moments in Adam, using $\beta_1 = 0.$, and $\beta_2 = 0.01$.
    \item \textit{Normalization matters.} This is more of an issue for SVGPs than exact GPs, but in both cases batchnorm can cause undesirable and unexpected behavior. Use layernorm.
\end{enumerate}

\paragraph{Applicable to variational GP inference}
\begin{enumerate}
    \item \textit{Initialization matters.} We use the procedure described in \citet{maddox2021conditioning} to reinitialize the inducing point locations and the variational parameters every time the model was retrained.
    This trick significantly improves results and saves computation, since the GP training does not completely start over every outer loop iteration.
    \item One final trick that is very useful for SVGPs is to turn off gradients to all GP-related parameters every other epoch (so half the epochs are only train the encoder). 
\end{enumerate}

\begin{figure}[h!]
    \centering
    \includegraphics[width=0.7\linewidth]{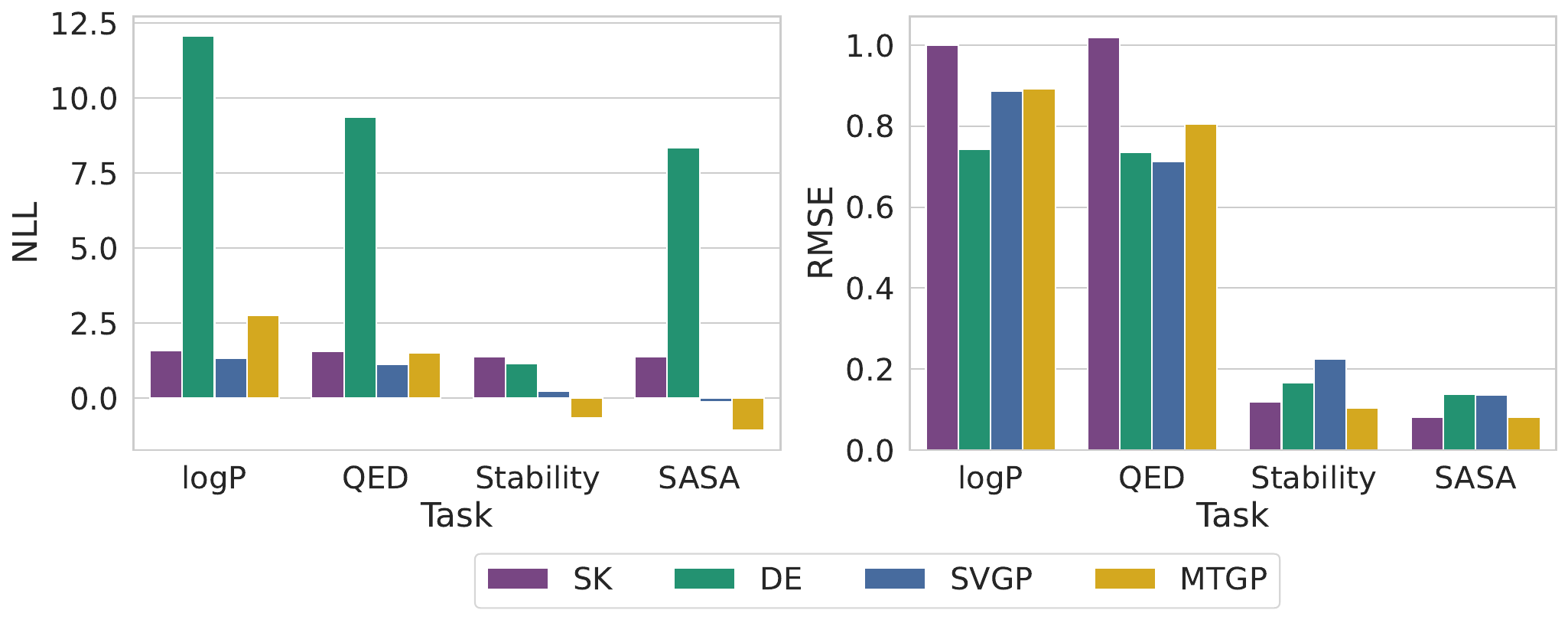}
    \caption{Negative log-likelihood (NLL) and root mean squared error (RMSE) for various discriminative models in the offline regression setting. DKL MTGPs and SVGPs have good performance across the board, while bootstrapped CNN ensembles (DE) are very overconfident. Exact GP inference with a substring kernel (SK) is very underconfident and has poor accuracy when predicting logP and QED.}
    \label{fig:surrogates_barplot}
\end{figure}

As we show in the main text and Figure \ref{fig:surrogates_barplot}, DKL SVGPs can consistently be trained to similar levels of accuracy as exact DKL GPs with very little trouble, once the proper training procedures are in place.
With these practical insights we were able to jointly train supervised GP heads and unsupervised language model heads on a shared encoder simply by taking one gradient step on the supervised GP loss and one gradient step on the unsupervised DAE loss per minibatch, using the same optimizer and learning rate schedule.
We used diagonal Gaussian likelihoods for all our experiments, with the noise variance initialized at $0.25$.

We found that DKL GPs (both exact and variational) were not immune to overfitting, so we used weight decay (1e-4) and reserved 10\% of all collected data (including online queries) as validation data for early stopping.

\subsection{DAE Implementation Details}

\paragraph{MLM Head}
We used a mask ratio of $0.125$ for all experiments when training MLM heads. The MLM loss is computed by randomly masking input tokens, and computing the empirical cross-entropy between the original sequence and the predictive distribution of the MLM head at the masked positions.
During sequence optimization the MLM predictive distribution is modified to prevent sampling the original token (to encourage diversity) and to prevent the sampling of special tokens.

\paragraph{LANMT Head}
Our LANMT head is identical to our MLM head, except for the addition of a length prediction head and length transform module \citep{shu2020latent}, a different corruption procedure and training objective. 
We used a max length change of $8$ in our experiments for Figure \ref{fig:lsbo_comparison}, so the corruption function randomly sampled a length change $\Delta t$ between -8 and 8. 
$\Delta t$ tokens were subsequently deleted, replaced, or inserted into the sequence.
The corrupted sequence was forwarded through the model, which was also given the original sequence length as a label during training. A training step takes a gradient step on the cross-entropy between the predicted length and the actual length, and on the cross-entropy between the predictive distribution over the whole decoded sequence and the original sequence.

\subsection{Hyperparameters}
\label{subsec:hypers}

\begin{table}[h!]
\centering
\parbox{.9\linewidth}{
\centering
\begin{tabular}{|l|c|}
\hline
\multicolumn{2}{|c|}{\textbf{Sequence Optimization}}\\
\hline
\textbf{Name} & \textbf{Value} \\
\hline
$|\mathcal{D}_0|$ & 512 \\
\hline
Query batch size ($b$) & 16 \\
\hline
$|\SeqBase|$ & $b$   \\
\hline
\# Optimization rounds ($i_{\mathrm{max}}$)                  & 64                        \\
\hline
\# Inner loop restarts                     & 16                    \\
\hline
\# Inner loop gradient steps ($j_{\mathrm{max}}$)                       & 32                        \\
\hline
Inner loop step size ($\eta$)          & 0.1 \\
\hline
Entropy penalty ($\lambda$)          & 1e-2 \\
\hline
\# MC acquisition samples & 2 \\
\hline
Random seeds & $\{0,\dots, 9\}$ \\
\hline
\end{tabular}
}\\
\vspace{4mm}
\parbox{.45\linewidth}{
\centering
\begin{tabular}{|l|c|}
\hline
\multicolumn{2}{|c|}{\textbf{DAE Architecture}}\\
\hline
\textbf{Name} & \textbf{Value} \\
\hline
Shared enc. depth (\# residual blocks) & 3 \\
\hline
Disc. enc. depth (\# residual blocks) & 1 \\
\hline
Decoder depth (\# residual blocks) & 3 \\
\hline
Conv. kernel width (\# tokens) & 5 \\
\hline
\# conv. channels & 64 \\
\hline
Latent dimension & 16 \\
\hline
GP likelihood variance init & 0.25 \\
\hline
GP lengthscale prior                  & $\mathcal{N}(0.7, 0.01)$                        \\
\hline
\# inducing points (SVGP head) & 64 \\
\hline
\end{tabular}
}
\hfill
\parbox{.45\linewidth}{
\centering
\begin{tabular}{|l|c|}
\hline
\multicolumn{2}{|c|}{\textbf{DAE Training}}\\
\hline
\textbf{Name} & \textbf{Value} \\
\hline
DAE corruption ratio (training) & 0.125 \\
\hline
DAE learning rate (MTGP head) & 5e-3 \\
\hline
DAE learning rate (SVGP head) & 1e-3   \\
\hline
DAE weight decay & 1e-4 \\
\hline
Adam EMA params ($\beta_1, \beta_2)$ & (0., 1e-2) \\
\hline
Early stopping holdout ratio                  & 0.1                        \\
\hline
Early stopping relative tolerance                     & 1e-3                    \\
\hline
Early stopping patience (\# epochs) & 32 \\
\hline
Max \# training epochs & 256 \\
\hline
\end{tabular}
}
\end{table}

\clearpage

\section{Additional Results}
\label{sec:additional_results}

\subsection{What About Substring Kernels?}
\label{subsec:ssk_offline_regression}

\begin{figure}[h!]
    \centering
\begin{subfigure}{0.2\textwidth}
\centering
\includegraphics[width=\linewidth]{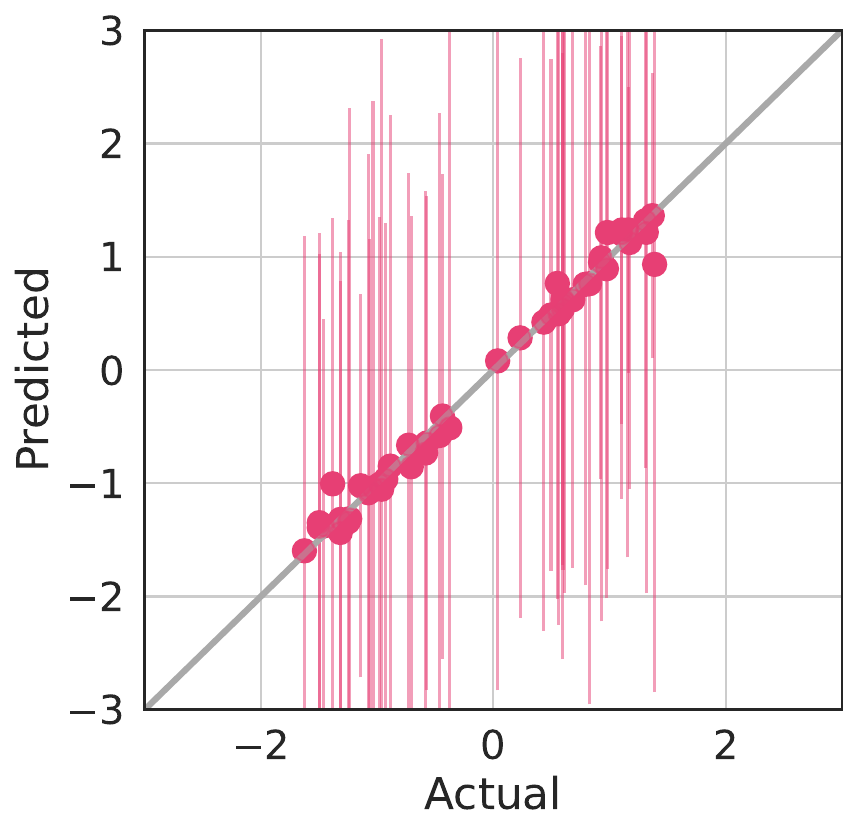}
\caption{SSK GP}
\end{subfigure}
\begin{subfigure}{0.2\textwidth}
\centering
\includegraphics[width=\linewidth]{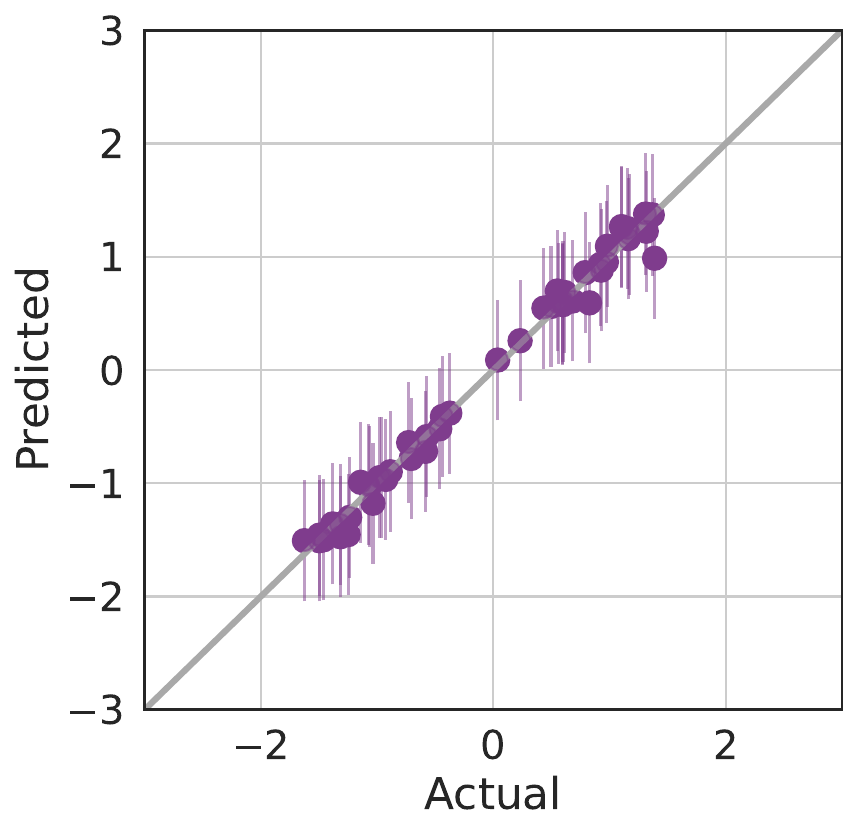}
\caption{DKL GP}
\end{subfigure}
    \caption{
    Evaluating the effect of kernel choice on exact GP regression with a discrete SSK \textbf{(left)} and a deep Mat\'ern kernel with a CNN encoder \textbf{(right)} when predicting the SASA property of RFP large molecules. 
    The SSK GP is under-confident and less accurate than the DKL GP, suggesting SSK GPs would not improve optimization performance.
    }
    \label{fig:surrogate_comp}
\end{figure}

Since the start pools used in the evaluation in Figure \ref{fig:hypervol_rel_comparison} are fairly small (i.e. 512 examples), it is natural to wonder how a substring kernel (SSK) GP \citep[e.g.,][]{moss2020boss}) would compete with our DKL-based GPs.
Due to constraints on time and computation, and SSK scalability issues, we do not directly compare SSK GPs and DKL GPs in the online setting.
However, in Figure \ref{fig:surrogate_comp}, we compare SSK GPs and DKL GPs in the offline regression setting,
training both models only through the supervised loss to predict the SASA property of RFP large molecules, using 410 examples for training and 102 examples for validation and test.
The SSK GP performs fairly well, but is very under-confident when compared to the DKL GP.
At an empirical level, our results do not support the claim that SSK GPs are superior models for biological sequences.
SSKs have further drawbacks which make it hard to justify additional investment into SSK GPs for drug design.

\textbf{Lack of Positional Information:} at a conceptual level, SSKs cannot identify regions of the sequence that have little effect on the objective values, since matching substrings increase the prior covariance between sequences regardless of where the substrings are found.
Simply put, an SSK just counts the occurrences of every possible $n$-gram across every possible combination of $n$ positions in a sequence.
The gap decay hyperparameter downweights occurrences corresponding to position combinations with elements that are not closely colocated.
Hence SSKs have very limited positional awareness in the sense that sequences with similar $n$-gram counts have high prior covariance, regardless of where the $n$-grams actually occurred.
Positional awareness is important when dealing with biological sequences from some subpopulation (e.g. a family of fluorescent proteins) since they have many identical subsequences, only varying at positions that strongly affect function.

\textbf{Difficult to Scale:} at a practical level, SSKs are difficult to integrate with deep generative models since they do not operate on continuous embeddings \citep{nigam2019augmenting}, and standard methods for scaling GPs --- inducing point methods \citep[e.g.,][]{hensman2013gaussian}, random feature expansions \citep[e.g.,][]{lazaro2010sparse}, and CG-based methods \citep[e.g.,][]{gardner2018gpytorch} --- are difficult to apply. 
In particular inducing points methods are impractical because the inducing point domain would be the discrete input space, introducing a challenging discrete optimization subproblem just to train the surrogate.
Furthermore SSKs struggle to scale not just to large datasets, but also to long sequences.
The dynamic programming algorithm used by \citet{moss2020boss} to compute their SSK is parallelizable, but becomes prohibitively memory intensive for sequences longer than 100 tokens, even when chunking the sequence into smaller pieces.
In fact, we used an Nvidia RTX 8000 GPU with 48 GB of memory just to produce Figure \ref{subsec:ssk_offline_regression}.
We also implemented a memory-efficient trie-based SSK, which could handle longer sequences but was prohibitively slow and difficult to parallelize.

\subsection{Comparison to LSBO for single-objective BBO}
\label{subsec:lsbo_comparison}

\begin{figure}[h!]
    \centering
    \includegraphics[height=0.24\textwidth]{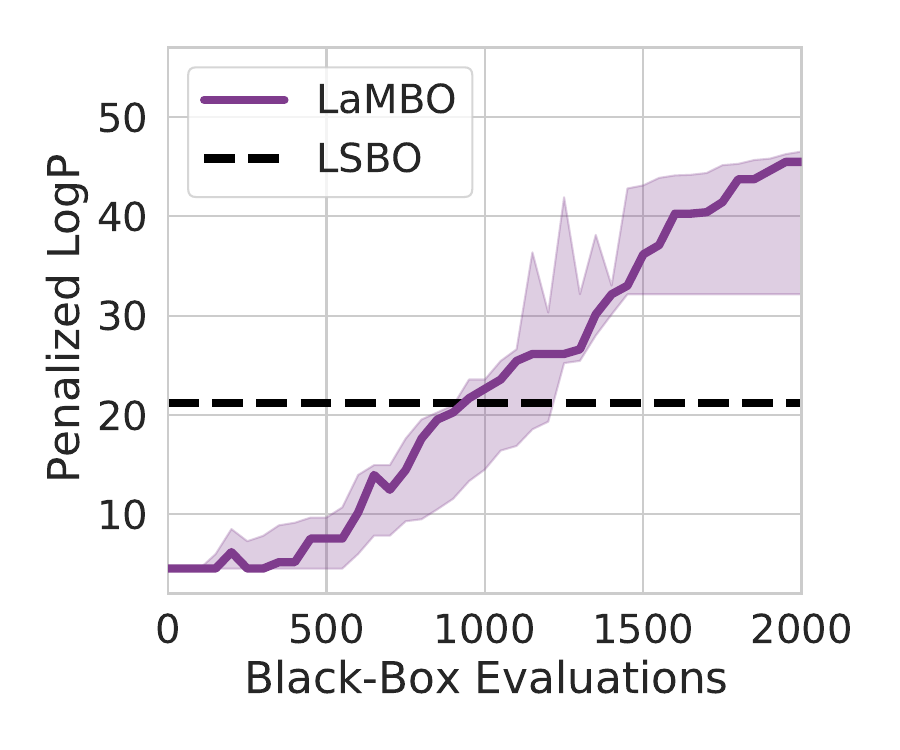}
    \caption{
    LaMBO reaches a higher objective value than the best reported LSBO result. When constrained to 500 evaluations LSBO is more sample efficient (ignoring pretraining data).
    The midpoint, lower, and upper bounds of the LaMBO curve depict the 60\%, 40\%, and 80\% quantiles, estimated from 5 trials.
    }
    \label{fig:lsbo_comparison}
\end{figure}

Here we evaluate LaMBO in the single-objective setting, using the \textit{penalized} logP task described in \citet{tripp2020sample}. 
In contrast to SSK-based methods, LaMBO can successfully be scaled to large datasets with standard variational GP inference, 
allowing us to compare to the popular latent space BayesOpt approach (LSBO) \citep{gomez2018automatic} on a larger-scale problem.
The start pool for this task is composed of the 2000 highest scoring sequences from ZINC, and 8000 random sequences, replicating the setup in \citet{tripp2020sample}. 
To accommodate the larger dataset, the discriminative head uses $k$ independent stochastic variational GPs (SVGPs) with 64 shared inducing points rather than an exact MTGP.

In Figure \ref{fig:lsbo_comparison} we demonstrate that LaMBO is competitive with a variant of LSBO specifically designed for this task, requiring about twice as many online observations before reaching the reported median best score attained by LSBO \citep{tripp2020sample}. 
As noted in Section \ref{sec:related_work}, LSBO uses the entire ZINC dataset for pretraining, so we do not directly compare sample efficiency.
In addition to the differences between LaMBO and LSBO already noted, we use SELFIES encodings rather than SMILES.
We use a seq2seq LANMT-style decoder head for this task, since logP heavily favors large molecules. High-scoring molecules such as those found by LSBO are larger than any found in ZINC, so it is important that the optimizer allow insertions.

Overall, LaMBO outperforms the best reported LSBO score (27.84) by a wide margin, reaching scores as high as 50 for some seeds, while using a more general architecture and requiring less data (counting both pretraining data and online queries). 
The factor that ultimately bounds the penalized logP objective in practice is the max sequence length constraint imposed by the positional encoding scheme we use. Therefore we note that, despite its widespread use, unconstrained logP (penalized or otherwise) is a poor optimization benchmark, since it can be manipulated by altering the positional encoding to permit longer sequences \citep{nigam2019augmenting,tripp2021fresh,fu2020mimosa}.

In short, while LaMBO is designed to facilitate multi-objective optimization --- a central feature of drug design --- 
it can also outperform the widely used single-objective LSBO, even in a single-objective setting. 

\clearpage
\subsection{Ablating LaMBO: Proposal Entropy and Discriminative Performance}

\begin{figure}[h!]
    \centering
    \includegraphics[width=0.8\textwidth,trim=0cm 0.6in 0cm 0.25in, clip]{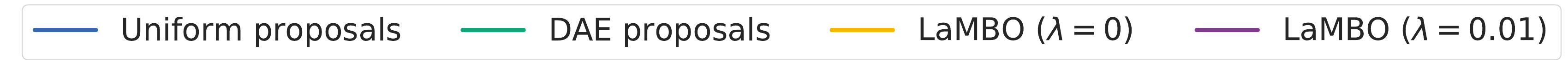}
    \\
    \begin{tabular}{cccc}
        \includegraphics[width=0.22\textwidth]{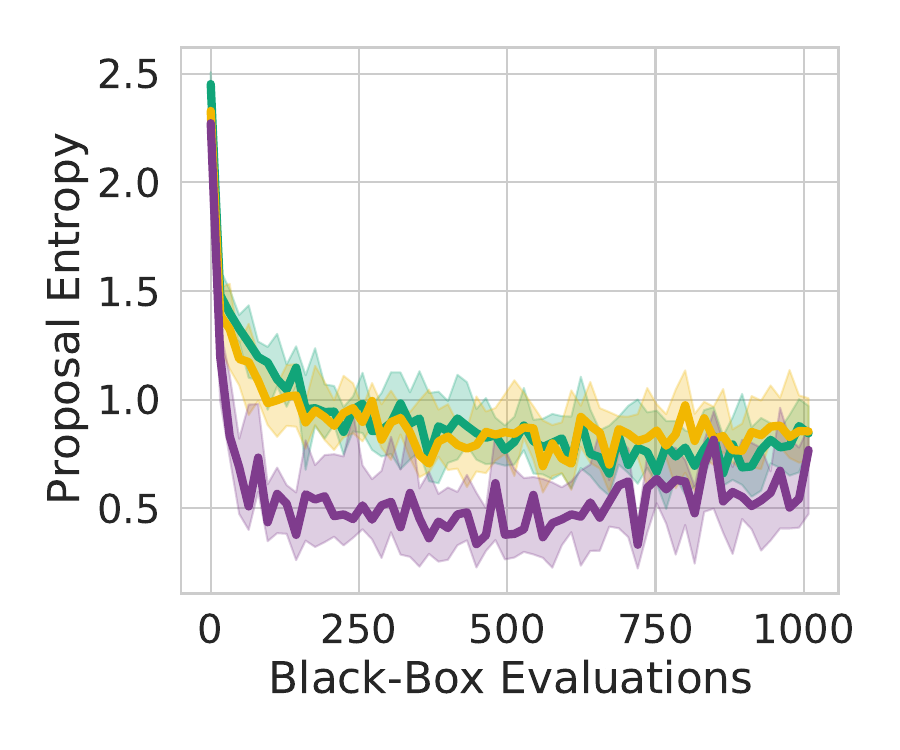}
        &
        \includegraphics[width=0.22\textwidth]{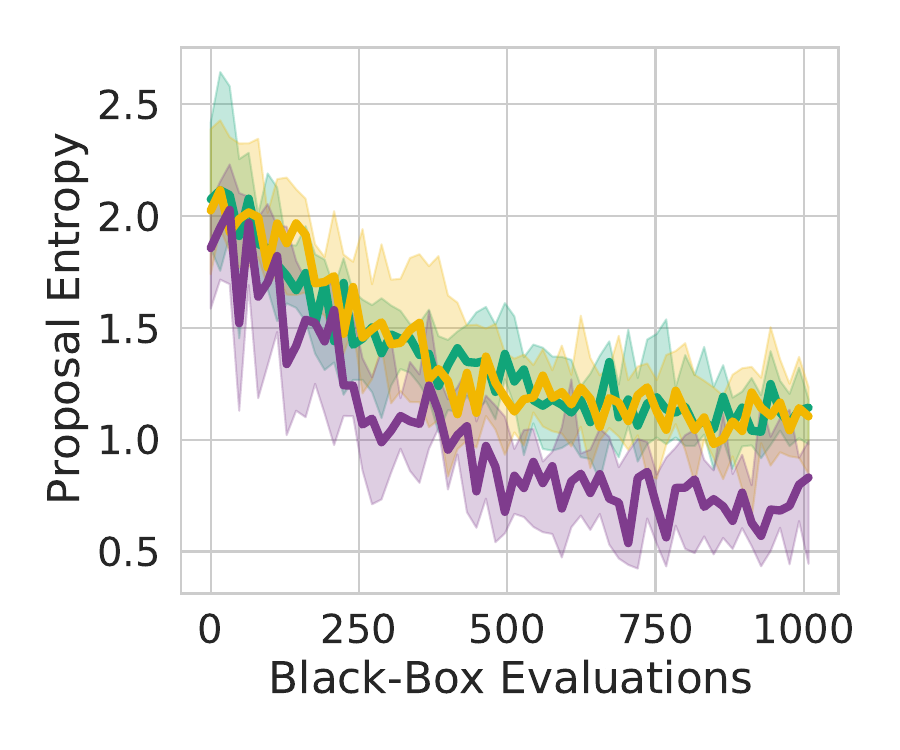}
        &
        \includegraphics[width=0.22\textwidth]{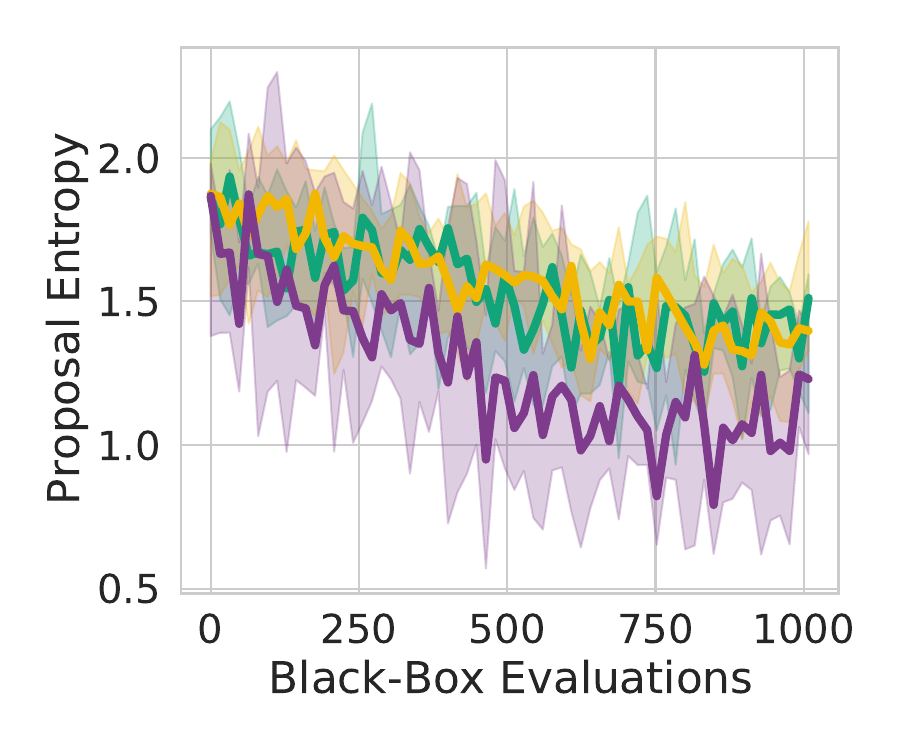}
        & 
        \includegraphics[width=0.22\textwidth]{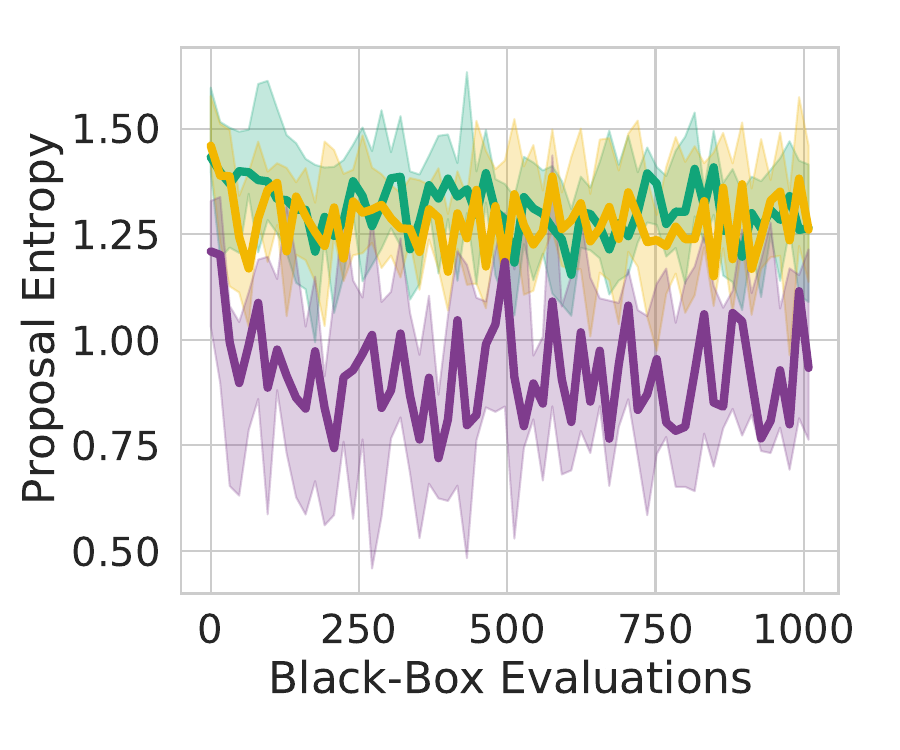} \\
        \includegraphics[width=0.22\textwidth]{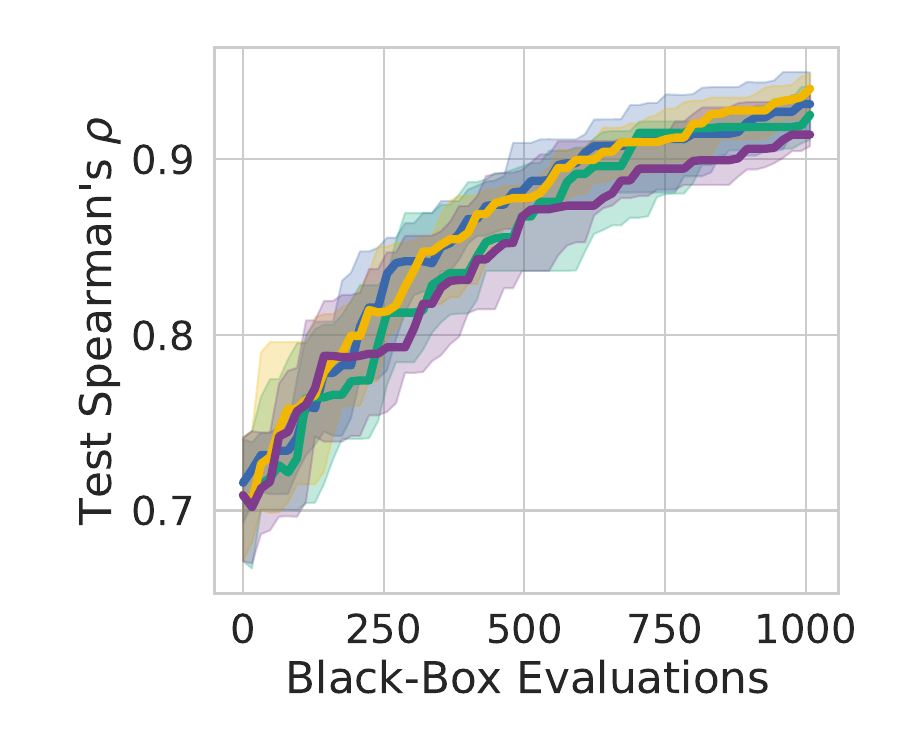}
        &
        \includegraphics[width=0.22\textwidth]{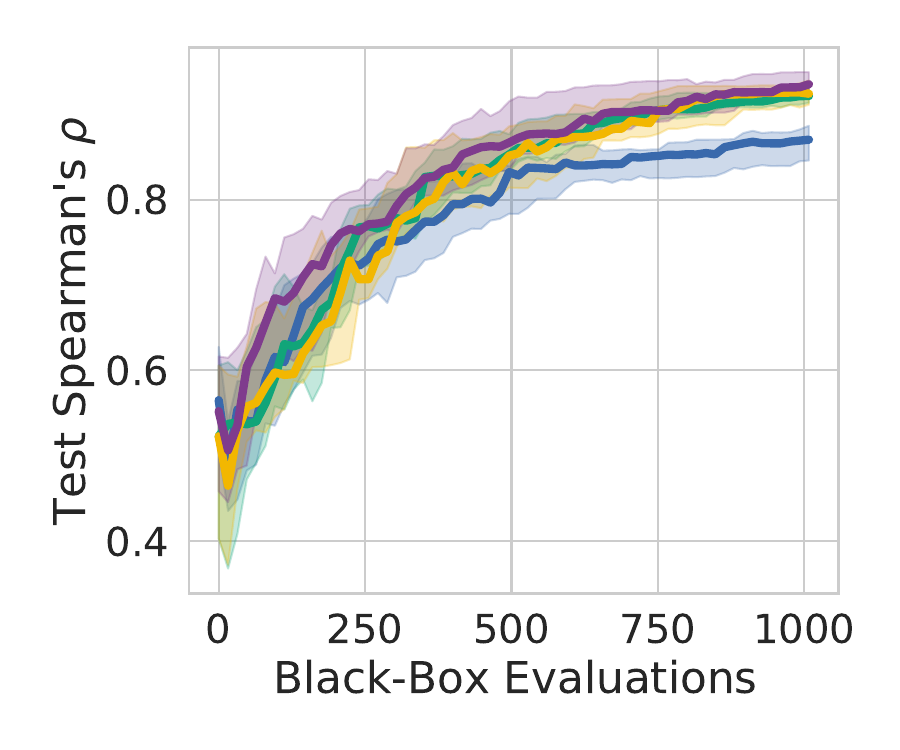}
        &
        \includegraphics[width=0.22\textwidth]{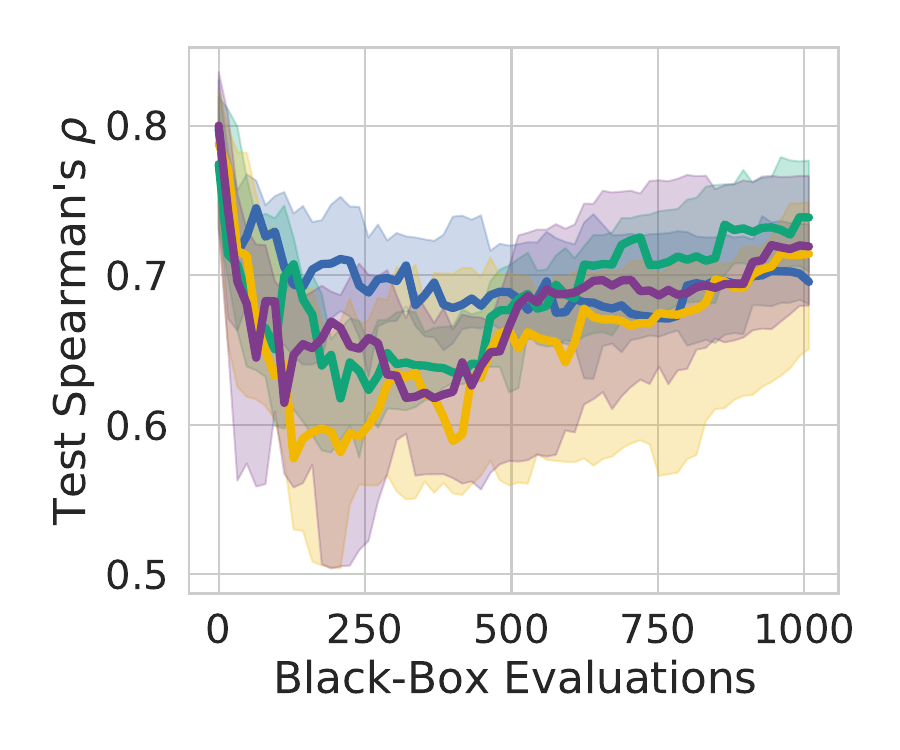}
        & 
        \includegraphics[width=0.22\textwidth]{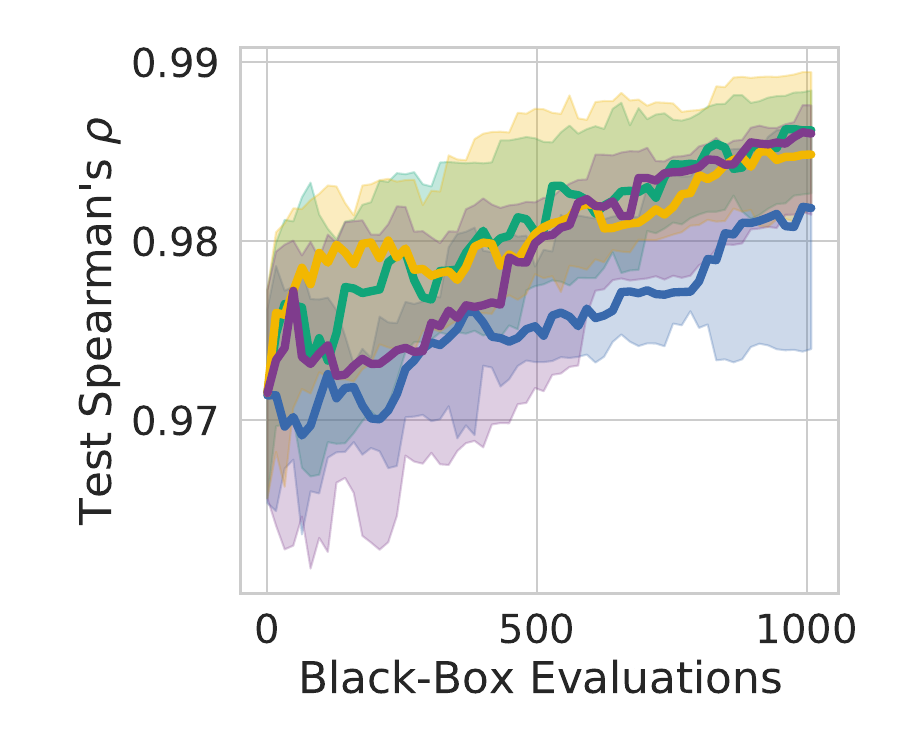} \\
        \includegraphics[width=0.22\textwidth]{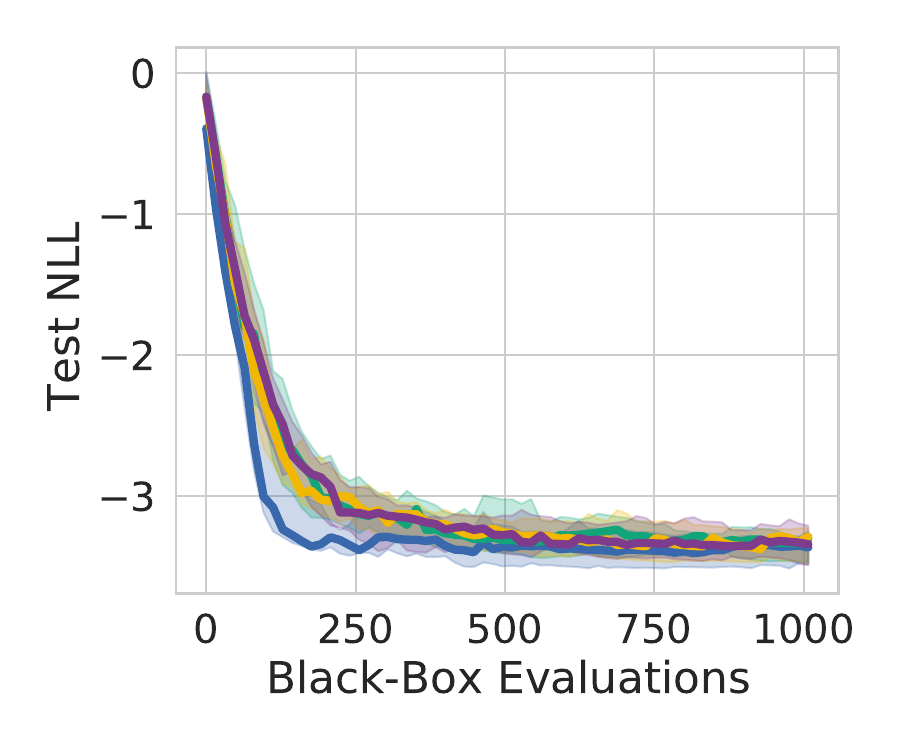}
        &
        \includegraphics[width=0.22\textwidth]{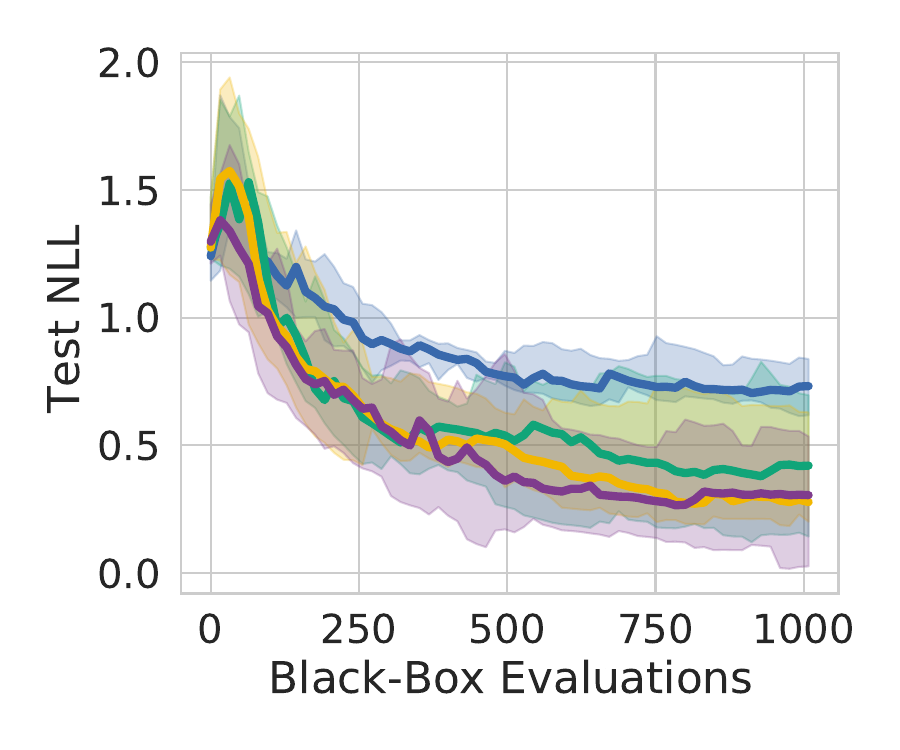}
        &
        \includegraphics[width=0.22\textwidth]{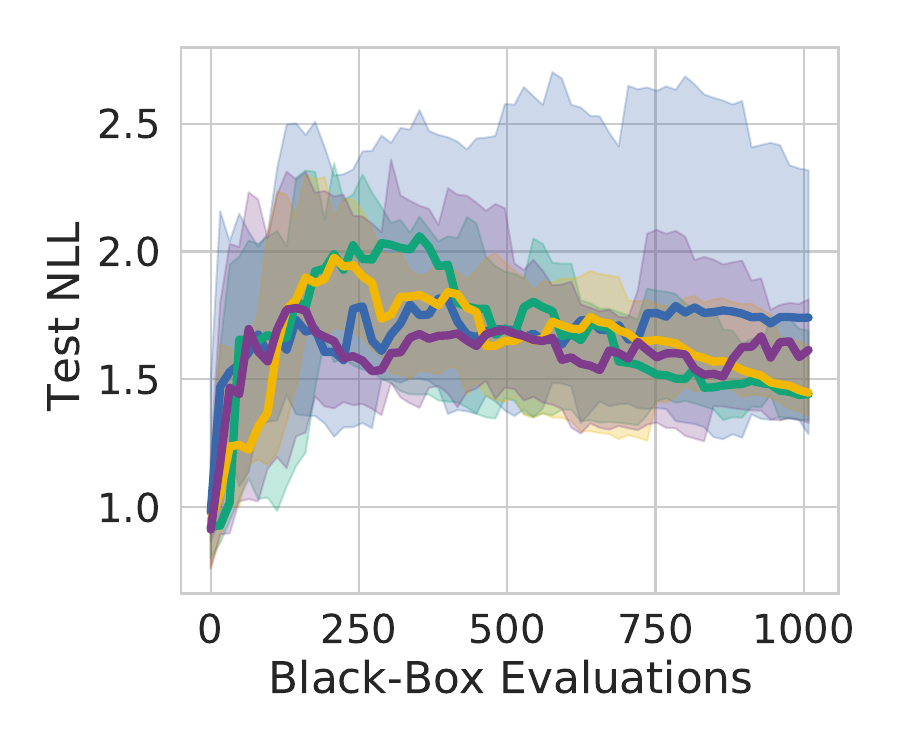}
        & 
        \includegraphics[width=0.22\textwidth]{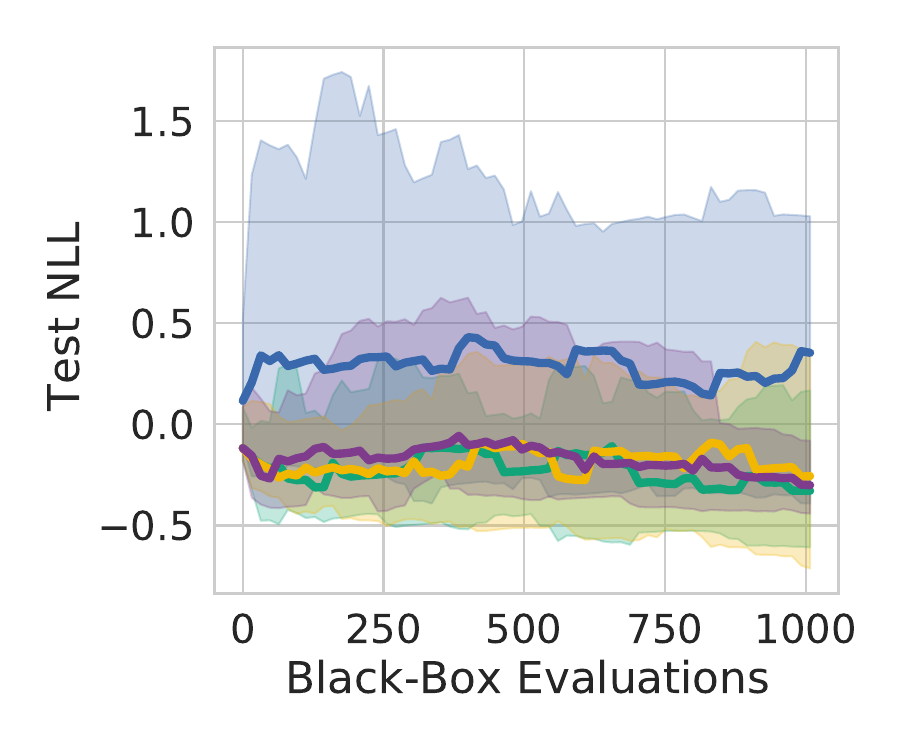} \\
        {\small \textbf{(a)} \textbf{Bigrams}} 
        & 
        {\small \textbf{(b)} \textbf{logP + QED}}
        & 
        {\small \textbf{(c)} \textbf{DRD3 docking + SA}}
        &
        {\small \textbf{(d)} \textbf{Stability + SASA}}
    \end{tabular}
    \caption{Additional metrics for the ablation experiment described in Section \ref{subsec:lambo_ablation} and Figure \ref{fig:lambo_ablation},
    where we cumulatively add the elements described in Section \ref{subsec:lambo_components}: (1) DAE-generated proposals, (2) DAE proposal optimization following $\nabla_Z[\ell_{\mathrm{query}}]$ with $\lambda = 0.$ (see Eq. \eqref{eq:reg_query_loss}), and (3) DAE proposal optimization with $\lambda = 0.01$.
    The \textbf{top} row shows the average entropy of the generative DAE proposal distributions over time.
    As expected, the entropy penalty decreases the proposal entropy.
    The \textbf{middle} and \textbf{bottom} rows show the discriminative Spearman's $\rho$ and NLL (averaged across objectives) on heldout data over time.
    Training the LaMBO architecture with both the unsupervised DAE objective and the supervised GP objective improves discriminative performance compared to the same model trained only through the supervised objective.
    Otherwise the methods behave similarly, verifying that better solutions are the result of better proposals, rather than a better discriminative model.
    The midpoint, lower, and upper bounds of each curve depict the 50\%, 20\%, and 80\% quantiles, estimated from 10 trials.
    }
    \label{fig:lambo_ablation_extras}
\end{figure}

\end{document}